\documentclass[10pt,twocolumn,letterpaper]{article}

\usepackage{cvpr}              
\usepackage{amsfonts}
\usepackage{amsmath}
\usepackage{bm}

\usepackage{algorithm}
\usepackage{algorithmicx}
\usepackage[accsupp]{axessibility}
\usepackage{algpseudocode}
\usepackage{color}
\usepackage[dvipsnames]{xcolor}
\usepackage{amsfonts} 
\usepackage{amsmath}
\usepackage{graphicx}
\usepackage{amssymb}
\usepackage{booktabs}
\usepackage{rotating}
\usepackage{multirow}
\usepackage{colortbl} 
\usepackage{tabulary}
\usepackage{etoolbox}
\usepackage{pifont}
\usepackage{makecell}
\usepackage[dvipsnames]{xcolor}

\definecolor{cvprblue}{rgb}{0.21,0.49,0.74}
\definecolor{mygray}{gray}{.9}
\usepackage[pagebackref,breaklinks,colorlinks,citecolor=cvprblue]{hyperref}

\def\paperID{6889} 
\def\confName{CVPR}
\def\confYear{2024}

\title{Domain-Agnostic Mutual Prompting for Unsupervised Domain Adaptation}

\author{Zhekai Du$^1$, Xinyao Li$^1$, Fengling Li$^2$, Ke Lu$^1$, Lei Zhu$^3$, Jingjing Li\thanks{Jingjing Li is the corresponding author.} $^1$\\
      $^1$University of Electronic Science and Technology of China; \\ $^2$University of Technology Sydney; $^3$Tongji University\\
      {\tt\small \{zhekaid, xinyao326\}@std.uestc.edu.cn, lijin117@yeah.net}
   }

\begin{document}
\maketitle
\begin{abstract}
 Conventional Unsupervised Domain Adaptation (UDA) strives to minimize distribution discrepancy between domains, which neglects to harness rich semantics from data and struggles to handle complex domain shifts. A promising technique is to leverage the knowledge of large-scale pre-trained vision-language models for more guided adaptation. Despite some endeavors, current methods often learn textual prompts to embed domain semantics for source and target domains separately and perform classification within each domain, limiting cross-domain knowledge transfer. Moreover, prompting only the language branch lacks flexibility to adapt both modalities dynamically. To bridge this gap, we propose Domain-Agnostic Mutual Prompting (DAMP) to exploit domain-invariant semantics by mutually aligning visual and textual embeddings. Specifically, the image contextual information is utilized to prompt the language branch in a domain-agnostic and instance-conditioned way. Meanwhile, visual prompts are imposed based on the domain-agnostic textual prompt to elicit domain-invariant visual embeddings. These two branches of prompts are learned mutually with a cross-attention module and regularized with a semantic-consistency loss and an instance-discrimination contrastive loss. Experiments on three UDA benchmarks demonstrate the superiority of DAMP over state-of-the-art approaches.
\end{abstract}    
\section{Introduction}
\label{sec:intro}

Labeling scarcity is a perennial problem in deep learning, as collecting abundant labeled data can be expensive, time-consuming, or even infeasible \cite{pan2009survey,zhang2023towards}. Unsupervised Domain Adaptation (UDA) serves as a promising approach to leverage the knowledge from a well-labeled source domain to benefit the task on an unlabeled target domain, where the two domains have similar semantics but different data distributions \cite{ganin2015unsupervised,tzeng2014deep,kang2019contrastive}. 

Conventional UDA methods typically bridge the domain gap by minimizing the distribution discrepancy, through either moment matching \cite{long2017deep,long2015learning,peng2019moment,kang2019contrastive,li2020maximum} or adversarial learning \cite{ganin2015unsupervised,long2018conditional,saito2018maximum,du2021cross}. However, roughly aligning two domains can result in distorted semantic structure and less class discriminability in learned feature representations \cite{tang2020unsupervised,chen2019transferability}. Besides, prior works use numerical labels for training and inference, which discard rich semantics behind categories, leading to sub-optimal adaptation when handling complex categories and domain shifts.
 
\begin{figure}[t]
  \centering
   \includegraphics[width=0.99\linewidth]{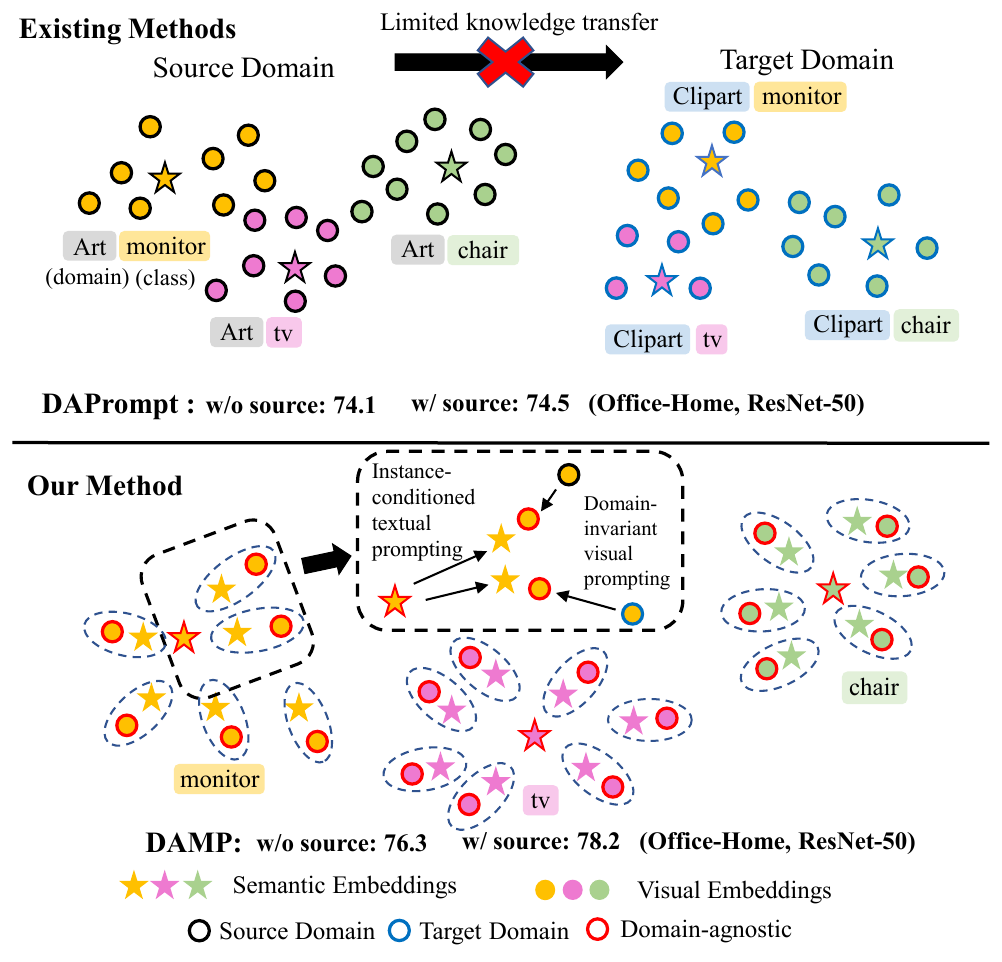}
   \vspace{-9pt}
   \caption{Top: exsiting prompt-based methods (e.g., DAPrompt \cite{ge2022domain}) only learn textual prompts to embed semantics for each domain and perform classification separately, which limits cross-domain knowledge transfer and feature alignment. Bottom: our method learns both textual and visual prompts mutually to make both modalities of embeddings domain-invariant, thus enabling better utilization of source knowledge and flexibility alignment.}
   \label{fig:illustration}
\vspace{-18pt}
\end{figure}

Recently, large-scale pre-trained Vision-Language Models (VLMs) have demonstrated impressive successes in various downstream tasks \cite{yu2023turning,jia2021scaling,rao2022denseclip,he2023clip}. By pre-training on tremendous image-text pairs, these models learn transferable multimodal representations that align images and texts in a joint embedding space. In particular, the Contrastive Language-Image Pre-training (CLIP) model \cite{radford2021learning} encodes rich semantic knowledge about visual concepts, presenting new opportunities to address the domain gap by leveraging the pre-trained vision and language knowledge. However, few attempts have been made to leverage VLMs for UDA since two challenges stand in the way, namely, 1) how to effectively take advantage of the rich {\it pre-trained knowledge} encoded in VLMs, and 2) how to transfer the {\it source knowledge} to the target domain for better adaptation.
 
Generally, there are two feasible routes for adapting large-scale pre-trained VLMs. The first is to use the zero-shot prediction capacity of VLMs to obtain pseudo-labels and fine-tune the image backbone with other UDA techniques \cite{lai2023padclip}. While the {\it source knowledge} can be well-encoded by fine-tuning, the pseudo-labels largely rely on manually designed textual descriptions and fine-tuning the model may ruin the {\it pre-trained knowledge}. Another way is to freeze the pre-trained model and only tune the input data (e.g., prompt) for model adaptation, which only involves a small set of learnable parameters and retains the {\it pre-trained knowledge}. For instance, DAPrompt \cite{ge2022domain} proposes to embed domain semantics into domain-specific textual prompts for each domain, which are then coupled with a domain-agnostic context for domain-specific classification in the joint CLIP space. However, we argue that a large portion of {\it source knowledge} is prone to be encoded in source-specific prompts, which cannot be transfered to the target domain. For instance, we conduct an experiment by disabling the source supervision loss in DAPrompt, and observe marginal influence on the target performance (see Fig. \ref{fig:illustration}). 

In this work, we aim to learn transferable (domain-agnostic) prompts to effectively leverage both {\it pre-trained knowledge} and {\it source-knowledge} for the target domain using CLIP. However, directly learning such textual prompts in UDA can be sub-optimal, as visual embeddings from different domains typically encompass distinct, domain-biased information that conforms to different distributions within the CLIP space. This is a key motivation behind domain-specific prompting in previous methods \cite{ge2022domain,singha2023ad}. Inspired by the recent success of visual prompting \cite{jia2022visual}, we propose also adapting the visual embeddings to elicit domain-invariant representations by prompting the vision backbone, based on the domain-agnostic textual prompt. Meanwhile, domain-invariant visual embeddings can still retain individual characteristics, e.g., object color and size. Such variations, even within the same category, necessitate instance-conditioned textual prompts for better alignment, as shown in Fig. \ref{fig:illustration}. Given the interdependent nature of the two kinds of prompts, we build a mutual learning framework based on a cross-attention mechanism inspired by the Transformer decoder \cite{vaswani2017attention}. A semantic-consistency regularization and an instance-discrimination contrastive loss are further imposed to ensure that the learned prompts carry pure domain-agnostic and instance-conditioned information. 

In summary, the key contributions of this work are three-fold: 1) We propose a novel framework termed DAMP to learn domain-agnostic prompts for transferring pre-trained knowledge and source knowledge to the target domain using CLIP. 2) DAMP mutually aligns textual and visual embeddings by prompting both modalities to learn domain-invariant representations, which are optimized with two elaborate regularizations. 3) Extensive experiments on three UDA benchmarks validate that DAMP brings consistent and notable gains over state-of-the-art approaches.
\section{Related Works}
\label{sec:rel_work}

{\bf Unsupervised Domain Adaptation.} To enable effective knowledge transfer, modern UDA methods typically fall into two technical routes. The first line of works aim to reduce the domain shift by aligning the feature distributions across domains. Common techniques include minimizing statistical distribution distances via moment matching \cite{long2017deep,peng2019moment,sun2016deep} and learning domain-invariant features via adversarial alignment \cite{ganin2015unsupervised,zhang2019domain,sankaranarayanan2018generate,saito2018maximum}. More recent methods focus on disentangling domain-invariant and domain-specific factors for casual invariance \cite{moraffah2019deep,yue2023make} or self-training with elaborate pseudo labels \cite{liu2021cycle,mei2020instance,wang2022debiased}. The second line of works resort to more large-scale networks, e.g., Vision Transformer (ViT) \cite{dosovitskiy2020image}, for more transferable features. For instance, CDTrans \cite{xu2021cdtrans} leverages the cross-attention mechanism in Transformer for cross-domain feature alignment. TVT \cite{yang2023tvt} introduces the evaluated transferabilities into the Multi-head Self-Attention module to construct a transferable ViT. SSRT \cite{sun2022safe} proposes to perturb the target features to refine the ViT and designs a safe training mechanism. 

Despite remarkable progresses, most existing UDA methods only operate in the vision modality, discarding the rich semantics behind features and categories, hindering effective adaptation for complex and large domain gaps.

{\bf Vision-Language Models and Prompt Learning.} Recent large-scale pre-trained VLMs have shown impressive performance on various vision-and-language tasks \cite{yao2021cpt,yu2023turning}. VLMs like CLIP \cite{radford2021learning} and ALIGN \cite{jia2021scaling} learn joint representations of images and texts by pre-training on large amounts of image-text pairs. A key capability of VLMs is the zero-shot prediction, where the pre-trained model can be applied to downstream tasks by simply conditioning on a textual prompt like ``a photo of a [CLS]''. This avoids costly fine-tuning and preserves the original knowledge in VLMs. However, manually designing effective prompts can be challenging. Prompt learning has thus become a popular VLMs adaptation technique. CoOp \cite{zhou2022learning} first uses learnable context tokens to prompt the language encoder of CLIP for visual classification. Later, CoCoOp \cite{zhou2022conditional} learns instance-conditioned prompts with a two-layer network for more generalizable textual prompts. MaPLe \cite{khattak2023maple} introduces multi-modal prompts to fine-tune both modalities. Neverthless, these works do not consider the domain shift problem. 

\begin{figure*}[t]
    \centering
     \includegraphics[width=0.99\linewidth]{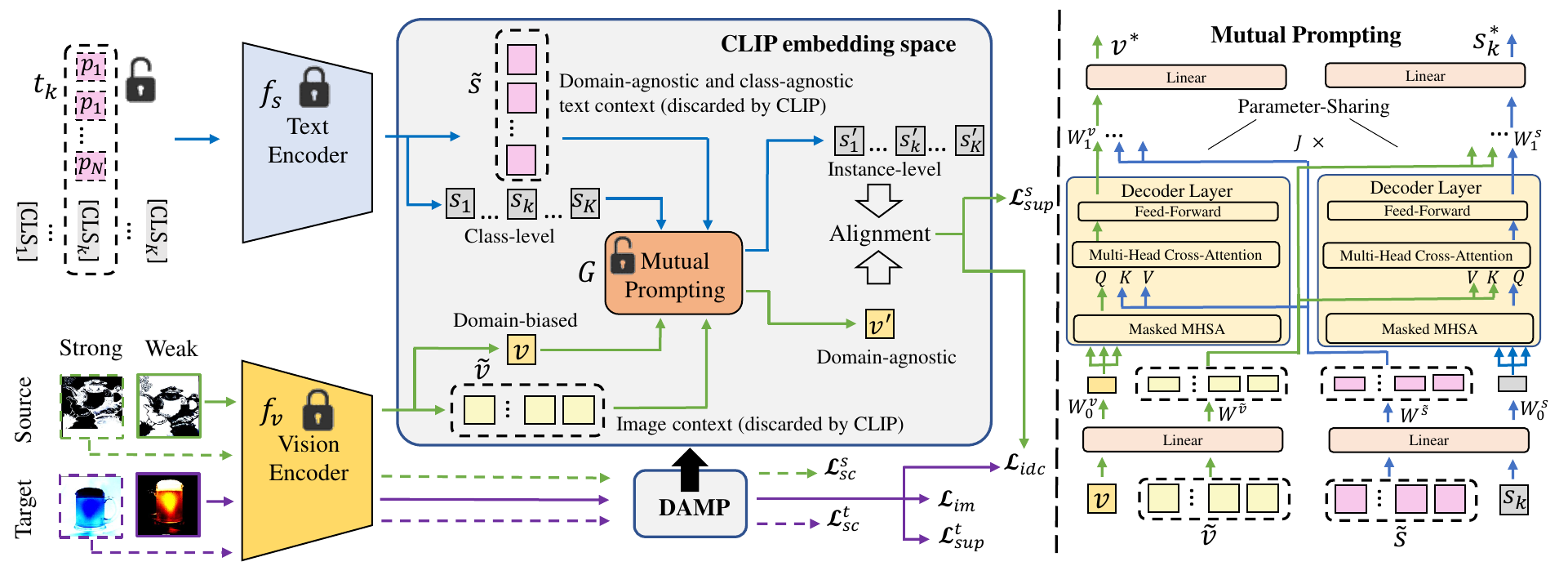}
     \vspace{-10pt}
     \caption{Overview of the proposed DAMP framework. Parameters of $f_s$ and $f_v$ are frozen and only $\bm{p}_{1:N}$ and $G$ are tunable during training. The blue arrows represent text data flows, while the green and purple arrows are data flows for source and target images, respectively. We only depict the prompting process for source weakly augmentated samples. All other samples follow the same process. $\mathcal{L}_{sc}^{s}$ ($\mathcal{L}_{sc}^{t}$), $\mathcal{L}_{idc}^{s}$ ($\mathcal{L}_{idc}^{t}$), and $\mathcal{L}_{im}$ are regularizations to make the prompting domain-agnostic, instance-conditioned and semantic-compatible, respectively.}
     \label{fig:framework}
  \vspace{-15pt}
  \end{figure*}

To leverage VLMs and prompt learning for UDA, DAPrompt \cite{ge2022domain} introduces a set of domain-specific textual tokens to encode domain semantics and perform classification with target-specific prompts. AD-CLIP \cite{singha2023ad} learns both domain- and image-specific tokens with feature statistics in the vision backbone. However, learning prompts for different domains separately may limit cross-domain knowledge transfer. Besides, prompting in a single modality cannot fully adapt the multi-modal knowledge in VLMs.

\section{Proposed Method}
\label{sec:method}
Let $\mathcal{D}_s = \{x^s_i, y^s_i\}_{i=1}^{N_s}$ be the source domain with $N_s$ labeled samples, where $x_s^i \sim P_s(X)$ is the input and $y_s^i \in Y$ is the label. Meanwhile, we have a target domain $\mathcal{D}_t = \{x_i^t\}_{i=1}^{N_t}$ with $N_t$ unlabeled samples and $x_t \sim P_t(X)$. These two domains are assumed to have different distributions in the data space $X$, but share the same label (semantic) set $Y$. The goal of UDA is to learn a model $f: X \rightarrow Y$ with $\mathcal{D}_s$ and $\mathcal{D}_t$ that can perform well on the target domain.

\subsection{Domain-Agnostic Prompting with CLIP}
Traditional UDA methods typically implement $f$ as a uni-modal neural network and associate each category with a numerical label, which overlook the rich semantics that could inform classification. In this work, we leverage CLIP \cite{radford2021learning} to enable semantic-driven classification.

CLIP learns aligned visual and textual representations by pre-training an image encoder $f_v$ and a text encoder $f_s$ on a large dataset of image-text pairs. Specifically, $f_v$ can be a ResNet \cite{he2016deep} or ViT \cite{dosovitskiy2020image} backbone that extracts a visual embedding from an image. On the other hand, $f_s$ uses a Transformer \cite{vaswani2017attention} to encode the paired textual description into a compact embedding. The two encoders are trained jointly with a contrastive loss. The aligned joint space then allows zero-shot classification for arbitrary input $\bm{x}$ \footnote{We use bold font $\bm{x}$ to denote either a source or a target sample.} by comparing the visual embedding $f_v(\bm{x})$ to textual embeddings $\{f_s(\bm{t}_k)\}_{k=1}^{K}$ correspond to $K$ classes in the joint space:
\begin{equation}
    P(\bm{y} = k \mid \bm{x})=\frac{\exp \left(cos\left(f_v(\bm{x}), f_s\left(\bm{t}_k\right)\right) / \tau\right)}{\sum_{k=1}^{K} \exp \left(cos\left(f_v(\bm{x}), f_t\left(\bm{t}_k\right)\right) / \tau\right)},
    \end{equation}
where $\tau=0.01$ is the temperature coefficient learned by CLIP, $cos(\cdot, \cdot)$ denotes the cosine similarity, and $\bm{t}_k$ is the textual prompt of the $k$-th class, e.g., ``a photo of a [CLS]''.

However, manually designed prompts can be naive and sub-optimal. A more effective way is to make $\{\bm{t}_k\}_{k=1}^K$ learnable as in CoOp \cite{{zhou2022learning}}. In UDA, covariate shift \cite{pan2009survey} is a widely adopted assumption, which indicates that the marginal distributions differ (i.e., $P_s(X) \neq P_t(X)$) but the conditional distribution $P(Y|X)$ remains unchanged between domains. This motivates using a shared set of textual prompts to model the invariant $P(Y|X)$. In this work, we use a domain- and class-shared input prompt template:
\begin{equation} \label{eq:ori_prompts}
\bm{t}_k := [\bm{p}_1] [\bm{p}_2] \dots [\bm{p}_N] [\text{CLS}_k],
\end{equation}
where $\bm{p}_{1:N} \in \mathbb{R}^{N \times D}$ are learnable contexts with length $N$ and dimension $D$, and $[\text{CLS}_k]$ is the $k$-th class name.

\subsection{Mutual Prompt Learning with Cross-Attention}
Directly learning domain-agnostic prompts as in Eq. \eqref{eq:ori_prompts} can be challenging in UDA. First, $f_v$ is pre-trained without domain adaptation objectives, yielding domain-biased visual embeddings that conform to different distributions across domains \cite{ge2022domain}. Second, the instance diversity leads to large intra-class variation, making it difficult to align all samples to a class-level textual prompt. To address these issues, we propose to impose visual prompts on $f_v$ to elicit more domain-agnostic visual representations. Meanwhile, we also adjust the textual prompt on $f_s$ according to each image contextual information for better image-text paired alignment, like in the original CLIP pre-training.

\textbf{Language-Guided Visual Prompting.} As $\{\bm{t}_k\}_{i=1}^{K}$ encode domain-agnostic class semantics, we can exploit these semantics to guide the generation of visual prompts that elicit domain-invariant visual characteristics. To achieve this, we use the cross-attention \cite{vaswani2017attention} mechanism to pass information between the two branches, which has shown great success in modeling multimodal interactions \cite{tsai2019multimodal,hu2021unit}. 

Given a textual prompt $\bm{t}_k$, the class name $[\text{CLS}_k]$ is first tokenized and embedded into $\bm{r}_k \in \mathbb{R}^{L_k \times D}$, where $L_k$ is the name length. The text encoder $f_s$ then extract embeddings via $J$ Transformer encoder layers $\{\texttt{Enc}_j\}_{j=1}^{J}$:
\begin{equation}
\begin{aligned}
    &T^k_j = \texttt{Enc}_j([\bm{p}_{1:N}, \bm{r}_k])~~~j=1. \\
    &T^k_j = \texttt{Enc}_j(T_{j-1})~~~j=2,3,\dots,J.
\end{aligned}
\end{equation}
Here $[\cdot, \cdot]$ stands for concatenation, $T^k_j \in \mathbb{R}^{(N+L_k) \times D}$ is the extracted embeddings in layer $j$. CLIP only uses the embedding at the last position of layer $J$ as the textual embedding, denoted as $\bm{s}_k$. However, we argue that embeddings at other positions also encode rich contextual information due to shared parameters among them. Therefore, we use the first $N$ embeddings of $T^k_J$, denoted as $\bm{\widetilde{s}} = T_J^k[1:N] \in \mathbb{R}^{N \times D}$, to guide the generation of visual prompts, which generally encode domain- and class-agnostic semantics.

For visual prompting, there are two widely used forms in the community, i.e., the pixel-level prompts \cite{bahng2022exploring,gan2023decorate} and token-level prompts \cite{jia2022visual,zha2023instance}. These two kinds of {\it pre-model} prompting poses challenges to prompt different vision backbones in a unified manner. In this work, we adopt the {\it post-model} prompting \cite{rao2022denseclip} strategy to prompt $f_v$ in the embedding space. Specifically, we first obtain the visual embedding ${\bm{v}}=f_v(\bm{x}) \in \mathbb{R}^{D}$ for an input $\bm{x}$, and aggregate information from text contexts $\bm{\widetilde{s}}$ via a cross-attention-based module $G$ with $L$ Transformer decoder layers $\{\texttt{Dec}_l\}_{l=1}^{L}$:
\begin{equation}
\begin{aligned} \label{eq:vision_update}
    [W^v_0, W^{\widetilde{s}}] &= \texttt{InProj}([\bm{v},~\bm{\widetilde{s}}]), \\
    W^v_l &= \texttt{Dec}_l(W^v_{l-1},W^{\widetilde{s}})~~~l=1,2,\dots,L, \\
    \bm{v}^* &= \texttt{OutProj}(W^v_L).
\end{aligned}
\end{equation}
Here $\texttt{InProj}$ and $\texttt{OutProj}$ are two projection operations, and each token is projected independently. We then obtain the final embedding $\bm{v}^{\prime}$  via a residual connection:
    $\bm{v}^{\prime} = \bm{v} + \gamma_v \bm{v}^*$,
where $\gamma_v$ controls the weight. This produces visual embeddings $\bm{v}^{\prime}$ guided by the domain-agnostic texts. 

\textbf{Vision-Guided Language Prompting.} To accommodate various visual backbones, CLIP uses a modified version of ResNet to implement $f_v$ by replacing the last Global Average Pooling (GAP) layer with an attention pooling layer. Specifically, it first transforms an image $\bm{x} \in \mathbb{R}^{H \times W \times 3}$ into a feature map $\bm{z} \in \mathbb{R}^{\hat{H} \times \hat{W} \times C}$, where $H (\hat{H}), W (\hat{W})$ and $C$ are the height, width and the channel number. The original ResNet uses $\bm{\overline{z}} = \texttt{GAP} (\bm{z}) \in \mathbb{R}^{C}$ as the final visual embedding. In CLIP, $\bm{z}$ and $\bm{\overline{z}}$ are further handled by a Multi-Head Self-Attention (MHSA) layer:
\begin{equation}
    \bm{v}, \bm{\widetilde{v}} = \texttt{MHSA}([\bm{\overline{z}}, \bm{z}]),
\end{equation}
where $\bm{v} \in \mathbb{R}^{1 \times D}$ and $\bm{\widetilde{v}} \in \mathbb{R}^{\hat{H} \hat{W} \times D}$ are the embeddings at the class token and other spatial positions, respectively, which are consistent with the ones in ViT. Generally, CLIP only uses $\bm{v}$ as the visual embedding and discards $\bm{\widetilde{v}}$. However, $\bm{\widetilde{v}}$ can also preserve useful semantical and spatial information that can be used as contextual information \cite{rao2022denseclip}. In this work, we leverage $\bm{\widetilde{v}}$ to adjust the textual embeddings $\{\bm{s}_k\}_{k=1}^{K}$ via the same {\it post-model} prompting strategy and $G$. Specifically, for the $k$-th class,
\begin{equation}
\begin{aligned} \label{eq:semantic_update}
    [W^s_0, W^{\widetilde{v}}] &= \texttt{InProj}([\bm{s}_k,\bm{\widetilde{v}}]), \\
    W^s_l &= \texttt{Dec}_l(W^s_{l-1},W^{\widetilde{v}})~~~l=1,2,\dots,L, \\
    \bm{s}^*_k &= \texttt{OutProj}(W^s_L).
\end{aligned}
\end{equation}
The final semantic embedding $\bm{s}_k^{\prime}$ is then obtained by:
    $\bm{s}_k^{\prime} = \bm{s}_k + \gamma_s \bm{s}_k^*$,
where $\gamma_s$ is weight coefficient for the text modality. Note that each $\bm{s}_k^{\prime}$ is updated based on a specific $\bm{x}$, making it instance-dependent, enabling better image-text alignment. These two branches of prompting are guided from each other to ensure mutual synergy. As a result, we use
\begin{equation}\label{eq:our_probability}
    \hat{P}(\bm{y} = k \mid \bm{x})=\frac{\exp \left(cos(\bm{v}^{\prime}, \bm{s}_{k}^{\prime}) / \tau\right)}{\sum_{k=1}^{K} \exp \left(cos(\bm{v}^{\prime}, \bm{s}_{k}^{\prime}) / \tau\right)}
\end{equation}
in our method for classification in the CLIP space.
\subsection{Auxiliary Regularizations}
While the mutual prompting framework aims to generate domain-invariant visual embeddings and instance-conditioned textual embeddings, directly optimizing with the source classification loss cannot guarantee achieving this goal. Hence, we design two auxiliary regularizations.

\textbf{Instance-Discrimination Contrastive Loss.} During the mutual prompting, the updated textual embeddings may still encode some domain-specific semantics from the image context, making $\{\bm{s}_{k}^{\prime}\}_{k=1}^K$ domain-biased and less capable for the target domain. To address this problem, we design an instance-discrimination contrastive loss to prevent textual prompts from learning domain-related cues from visual contexts. Our motivation is that images from the same domain typically share the same domain-information. Therefore, maximizing the difference in $\{\bm{s}_k^{\prime}\}_{k=1}^K$ among them help remove the domain-specific information. 

Specifically, given a batch of source or target samples $\mathcal{B}$, denote $\{\bm{s}_{a,k}^{\prime}\}_{k=1}^K$ and $\bm{v}_a^{\prime}$ the textual and visual embeddings after mutual prompting for $\bm{x}_a \sim \mathcal{B}$, each $\bm{x}_a$ forms a positive pair for $\bm{v}_a^{\prime}$ and $\{\bm{s}_{a,k}^{\prime}\}_{k=1}^K$ and forms negative pairs for $\{\bm{s}_{a,k}^{\prime}\}_{k=1}^K$ and $\bm{v}_b^{\prime}$ from another $\bm{x}_b$ within the same batch:
\begin{equation} \label{eq:ind}
  \begin{aligned}
    & \operatorname{sim}(\bm{x}_a, \bm{x}_b) = \frac{1}{K} \sum_{k=1}^{K} cos(\bm{s}_{a,k}^{\prime}, \bm{v}_b^{\prime}) / \tau, \\
    & \mathcal{L}_{idc} = -\log \frac{\exp \left(\operatorname{sim}(\bm{x}_a, \bm{x}_a)\right)}{\operatorname{sim}(\bm{x}_a, \bm{x}_a) + \sum_{b \neq a} \operatorname{sim}(\bm{x}_a, \bm{x}_b)}.
  \end{aligned}
\end{equation}
This contrastive loss forces $\{\bm{s}_{k}^{\prime}\}_{k=1}^K$ to not encode domain-specific cues while retaining pure instance-specific information. Imagine that if $\{\bm{s}_{k}^{\prime}\}_{k=1}^K$ contained domain-related information, they would be more similar for different images from the same domain, thus the domain-specific information can be further removed by optimizing $\mathcal{L}_{idc}$. Meanwhile, this contrastive loss can be optimized in an unsupervised way, thus providing regularizations for both domains.

\textbf{Semantic-Consistency Regularization.} In addition to removing domain-specific information in $\{\bm{s}_{k}^{\prime}\}_{k=1}^K$, we also want to ensure the prompted visual embedding $\bm{v}^{\prime}$ is domain-invariant. Inspired by FixMatch \cite{sohn2020fixmatch}, we aim to exploit domain-agnostic visual characteristics with a semantic-consistency regularization. Concretely, we leverage RandAugment \cite{cubuk2020randaugment} to obtain a strongly-augmentated version of $\bm{x}$, denoted as $\mathcal{A}(\bm{x})$, and enforce it to be correctly classified. For labeled source samples $\{x^s_i, y^s_i\}_{i=1}^{N_s}$, we can directly optimize with ground-truth labels via:
\begin{equation} \label{eq:source_sc}
  \mathcal{L}^s_{sc} = - \sum_{i=1}^{N_s} \log \hat{P}(\bm{y} = y^s_i \mid \mathcal{A}(x^s_i)).
\end{equation}
Meanwhile, we also obtain pseudo-labels $\{\hat{y^t_i}\}_{i=1}^{N_t}$ for target data $\{x_i^t\}_{i=1}^{N_t}$ and only involves confident ones for training:
\begin{equation} \label{eq:target_sc}
    \mathcal{L}^t_{sc} = - \sum_{i=1}^{N_t} \mathbb{I}\{\hat{P}(\bm{y} = \hat{y^t_i} \mid S(x^t_i)) \ge T\} \log \hat{P}(\bm{y} = \hat{y^t_i} \mid \mathcal{A}(x^t_i)),
\end{equation}
where $\mathbb{I}{\cdot}$ is an indication function, $T$ is the threshold for filtering confident target samples.

However, the unconfident target sampels are still not well-exploited. To make the updated target domain embeddings fit the learned semantic structure, we leverage the information maximization \cite{liang2020we, hu2017learning} technique to regularize the unlabeled target data via an entropy-based loss:
\begin{equation} \label{eq:im}
    \mathcal{L}_{im}= \frac{1}{N_{t}} \sum_{i=1}^{N_{t}} \sum_{k=1}^{K} p_{t i}^{c} \log p_{t i}^{c} - \sum_{k=1}^{K} \hat{p}_t^{k} \log \hat{p}_t^{k},
\end{equation}
where $p_{ti}^k = \hat{P}(\bm{y}=k \mid x^t_i)$ and $\hat{p}_t^k=\frac{1}{N_{t}} \sum_{j=1}^{N_{t}} p_{t j}^k$. Optimizing $\mathcal{L}_{im}$ makes predictions globally diverse and locally confident, thus avoiding category collapse and ambiguity.

\subsection{Overall Training Objective}
We train our DAMP with the supervised loss and above regularizations in an end-to-end manner. For the source domain, the supervised loss can be expressed by:
\begin{equation} \label{eq:source_sup}
    \mathcal{L}^s_{sup} = - \sum_{i=1}^{N_s} \log \hat{P}(\bm{y} = y^s_i \mid \mathcal{W}(x^s_i)),
  \end{equation}
where $\mathcal{W}(\cdot)$ is a weak augmentation operation. Besides, we also supervise target samples with confident pseudo-labels:
\begin{equation} \label{eq:target_sup}
    \mathcal{L}^t_{sup} = - \sum_{i=1}^{N_t} \mathbb{I}\{\hat{P}(\bm{y} = \hat{y^t_i} \mid \mathcal{W}(x^t_i)) \ge T\} \log \hat{P}(\bm{y} = \hat{y^t_i} \mid \mathcal{W}(x^t_i)).
\end{equation}
The final training objective $\mathcal{L}_{all}$ is formulated by
\begin{equation}
    \mathcal{L}_{all} = \mathcal{L}_{sup} + \mathcal{L}_{sc} + \lambda_{c} \mathcal{L}_{idc} + \lambda_{i} \mathcal{L}_{im}, 
\end{equation}
where $\lambda_{c}$ and $\lambda_{i}$ are trade-off weights, $\mathcal{L}_{sc}=\mathcal{L}^s_{sc}+\mathcal{L}^t_{sc}$ and $\mathcal{L}_{sup}=\mathcal{L}^s_{sup}+\mathcal{L}^t_{sup}$. We give equal weights to $\mathcal{L}_{sup}$ and $\mathcal{L}_{sc}$ for treating different augmentations equally. An overview of our method can be found in Fig. \ref{fig:framework}.
\section{Experiments}
\label{sec:experiments} 

\begin{table*}[t]
        \centering
        \caption{Classification accuracies (\%) on {\bf Office-Home} dataset for UDA. The best and second best results within each backbone are highlighted in bold and underline, respectively. $\dagger$ CDTrans uses DeiT-B \cite{touvron2021training} as the backbone. Methods within each backbone are grouped into three categories, i.e., fine-tuning, zero-shot and prompt learning (from top to bottom), respectively.}
        \vspace{-5pt}
        \resizebox{0.95\linewidth}{!}{
        \begin{tabular}{l|c|cccccccccccccc} 
        \toprule
           {\textbf{Method}} & $f_v$ & {Ar$\rightarrow$Cl} & {Ar$\rightarrow$Pr} & {Ar$\rightarrow$Rw} & {Cl$\rightarrow$Ar} & {Cl$\rightarrow$Pr} & {Cl$\rightarrow$Rw} & {Pr$\rightarrow$Ar} & {Pr$\rightarrow$Cl} & {Pr$\rightarrow$Rw} & {Rw$\rightarrow$Ar} & {Rw$\rightarrow$Cl} & {Rw$\rightarrow$Pr} & {\textbf{Avg.}} \\ 
            \midrule
            ResNet-50 \cite{he2016deep} & & 34.9& 50.0& 58.0& 37.4& 41.9& 46.2& 38.5& 31.2& 60.4& 53.9& 41.2& 59.9& 46.1  \\
    
    
    
    
            SRDC \cite{tang2020unsupervised}& & 52.3& 76.3& 81.0& 69.5& 76.2& 78.0& 68.7& 53.8& 81.7& 76.3& 57.1& 85.0& 71.3 \\

            ToAlign \cite{wei2021toalign}& & 57.9 & 76.9 & 80.8 & 66.7 & 75.6 & 77.0 & 67.8 & 57.0 & 82.5 & 75.1 & 60.0 & 84.9 & 72.0 \\
    
            + FixMatch + EIDCo \cite{zhang2023towards}& & {\bf 63.8} & 80.8 & 82.6 & 71.5 & 80.1 & 80.9 & 72.1 & {\bf 61.3} & 84.5 & {\bf 78.6} & {\bf 65.8} &  \underline{87.1} & 75.8 \\

            PADCLIP \cite{lai2023padclip}& & 57.5 & 84.0 & 83.8 & {\bf 77.8} & 85.5 & 84.7 & \underline{76.3} & 59.2 & \underline{85.4} & \underline{78.1} & 60.2 & 86.7 & \underline{76.6} \\
            \cmidrule{1-1} \cmidrule{3-15}

            CLIP \cite{radford2021learning}& & 51.6& 81.9& 82.6& 71.9& 81.9& 82.6& 71.9& 51.6& 82.6& 71.9& 51.6& 81.9& 72.0 \\
            \cmidrule{1-1} \cmidrule{3-15}
            DAPrompt \cite{ge2022domain}& & 54.1& 84.3& 84.8& 74.4& 83.7& 85.0& 74.5& 54.6& 84.8& 75.2& 54.7& 83.8& 74.5 \\

            AD-CLIP \cite{singha2023ad} & & \multirowcell{-8}[-5.0ex]{\hspace*{-1.8em} \vspace{2em}\turnbox{90}{\thead{ResNet-50}}} \hspace{-1.0cm} 55.4& \underline{85.2}& \underline{85.6}& 76.1& \underline{85.8}& \underline{86.2}& {\bf 76.7}& 56.1& \underline{85.4}& 76.8& 56.1& 85.5& 75.9 \\       

            \rowcolor{mygray} DAMP (Ours) & & \underline{59.7} & {\bf 88.5} & {\bf 86.8} & \underline{76.6} & {\bf 88.9} & {\bf 87.0} &	\underline{76.3} & \underline{59.6} & {\bf 87.1} & 77.0 & \underline{61.0} & {\bf 89.9} & \textbf{78.2} \\
    \midrule
    \midrule    
            ViT-B \cite{dosovitskiy2020image} & &  54.7 & 83.0 & 87.2 & 77.3 & 83.4 & 85.5 & 74.4 & 50.9 & 87.2 & 79.6 & 53.8 & 88.8 & 75.5 \\        

            CDTrans $\dagger$ \cite{xu2021cdtrans}& & 68.8 & 85.0& 86.9& 81.5& 87.1& 87.3& 79.6& 63.3& 88.2& 82.0& 66.0& 90.6& 80.5\\
    
            TVT-B \cite{yang2023tvt}& & 74.9& 86.8& 89.5& 82.8& 88.0& 88.3& 79.8& 71.9& 90.1& 85.5& 74.6& 90.6& 83.6\\
    
            SSRT-B \cite{sun2022safe}& & 75.2& 89.0& 91.1& 85.1& 88.3& 90.0& 85.0& 74.2& 91.3& 85.7& 78.6& 91.8& 85.4\\

            + FixMatch + EIDCo \cite{zhang2023towards}& &  {\bf 76.9} & 90.3 & 91.3 & \underline{86.5} & 90.5 & 90.0 & \underline{86.3} & \underline{75.5} & 91.7 & {\bf 88.1} & \underline{77.1} & 92.3 & 86.4 \\
            PADCLIP \cite{lai2023padclip} & & \underline{76.4} & 90.6 & 90.8 & {\bf 86.7} & 92.3 & \underline{92.0} & 86.0 & 74.5 & 91.5 & \underline{86.9} & {\bf 79.1} & 93.1 & \underline{86.7} \\
            \cmidrule{1-1} \cmidrule{3-15}
            CLIP \cite{radford2021learning} & & 67.8& 89.0& 89.8 & 82.9& 89.0& 89.8& 82.9& 67.8& 89.8& 82.9& 67.8& 89.0& 82.4  \\
            \cmidrule{1-1} \cmidrule{3-15}
            DAPrompt \cite{ge2022domain} & & \multirowcell{-6}[-4.8ex]{\hspace*{-2.8em} \vspace{1.5em}\turnbox{90}{\thead{ViT-B/16}}} \hspace{-1.0cm} 70.7& 91.0& 90.9& 85.2& 91.0& 91.0& 85.1& 70.7& 90.9& 85.3& 70.4& 91.4& 84.4 \\ 
    
            AD-CLIP \cite{singha2023ad} & & 70.9& \underline{92.5}& {\bf 92.1}& 85.4& \underline{92.4}& {\bf 92.5}& {\bf 86.7}& 74.3& {\bf 93.0}& \underline{86.9} & 72.6& \underline{93.8}& 86.1 \\          
            \rowcolor{mygray} DAMP (Ours) & & 75.7 & {\bf 94.2} & \underline{92.0} & 86.3 & {\bf 94.2} & 91.9 & 86.2 & {\bf 76.3} & \underline{92.4} & 86.1 & 75.6 & {\bf 94.0} & \textbf{87.1} \\ 
        \bottomrule
        \end{tabular}}
        \vspace{-5pt}
        \label{tab:officehome}
    \end{table*}

    \begin{table*}[t]
        \centering
        \caption{Per-class accuracies (\%) on {\bf VisDA-17} dataset for UDA. Marks and symbols share the same meaning in Table \ref{tab:officehome}. }\label{tab:visda}
        \vspace{-5pt}
        \scriptsize
        \setlength{\tabcolsep}{2.6mm}{
        \begin{tabular}{l|c|cccccccccccccc}
        \toprule
        {\textbf{Method}} & $f_v$ & {plane} & {bicycle} & {bus} & {car} & {horse} & {knife} & {mcycl} & {person} & {plant} & {sktbrd} & {train} & {truck} & {Avg} \\ 
            \midrule
    
            RN-101 \cite{he2016deep} & & 55.1& 53.3& 61.9& 59.1& 80.6& 17.9& 79.7& 31.2& 81.0& 26.5& 73.5& 8.5& 52.4 \\

            CGDM \cite{du2021cross} & &  93.4 & 82.7 & 73.2 & 68.4 & 92.9 & 94.5 & 88.7 & 82.1 & 93.4 & 82.5 & 86.8 & 49.2 & 82.3\\

            CAN \cite{kang2019contrastive} & & 97.0 & 87.2 & 82.5 & 74.3& \underline{97.8}& {\bf 96.2}& 90.8& 80.7& {\bf 96.6} & {\bf 96.3} &87.5& 59.9& 87.2\\

            PADCLIP \cite{lai2023padclip} & & 96.7 & 88.8 & 87.0 & {\bf 82.8} & 97.1 & 93.0 & 91.3 & \underline{83.0} & \underline{95.5} & 91.8 & 91.5 & 63.0 & {\bf 88.5}  \\
            \cmidrule{1-1} \cmidrule{3-15}
            CLIP \cite{radford2021learning} & & 98.2& 83.9& \underline{90.5}& 73.5& 97.2& 84.0& \underline{95.3}& 65.7& 79.4& 89.9& 91.8& 63.3& 84.4  \\
            \cmidrule{1-1} \cmidrule{3-15}
    
            DAPrompt \cite{ge2022domain} & & \underline{97.8}& 83.1& 88.8& \underline{77.9}& 97.4& 91.5& 94.2& 79.7& 88.6& 89.3& {\bf 92.5}& 62.0&  86.9  \\

            AD-CLIP \cite{singha2023ad} & & \multirowcell{-8}[-5.0ex]{\hspace*{-1.7em} \vspace{-1.2em}\turnbox{90}{\thead{ResNet-101}}} \hspace{-0.36cm} {\bf 98.1}& 83.6& {\bf 91.2}& 76.6& {\bf 98.1}& 93.4& \bf{96.0}& 81.4& 86.4& 91.5& \underline{92.1}& 64.2& 87.7  \\

            \rowcolor{mygray} DAMP (Ours) & & 97.3& {\bf 91.6}& 89.1& 76.4& 97.5& 94.0& 92.3& {\bf 84.5}& 91.2& 88.1& 91.2& {\bf 67.0}& \underline{88.4} \\
            \midrule
            \midrule
            ViT-B \cite{dosovitskiy2020image}& & 99.1 & 60.7 & 70.6 & 82.7 & 96.5 & 73.1 & {\bf 97.1} & 19.7 & 64.5 & 94.7 & {\bf 97.2} & 15.4 & 72.6\\

            CDTrans $\dagger$ \cite{xu2021cdtrans}& & 97.1& 90.5& 82.4& 77.5& 96.6& 96.1& 93.6& \textbf{88.6}& \textbf{97.9}& 86.9& 90.3& 62.8& 88.4 \\ 
    
            TVT-B \cite{yang2023tvt}& &  92.9 & 85.6 &77.5 &60.5& 93.6& \underline{98.2}& 89.4& 76.4 &93.6 &92.0 &91.7 &55.7& 83.9\\
    
            SSRT-B \cite{sun2022safe}& & 98.9& 87.6& 89.1& \underline{84.8}& 98.3& \textbf{98.7}& 96.3& 81.1& \underline{94.9}& {\bf 97.9}& 94.5& 43.1& 88.8\\
            
            PADCLIP \cite{lai2023padclip} & & 98.1 & {\bf 93.8}& 87.1& {\bf 85.5} &98.0& 96.0& 94.4& \underline{86.0}& \underline{94.9}& 93.3 &93.5& {\bf 70.2}& {\bf 90.9} \\ 
            \cmidrule{1-1} \cmidrule{3-15}
            CLIP \cite{radford2021learning} & & 99.1& 91.7& \underline{93.8} & 76.7& 98.4& 91.7& 95.3& 82.7& 86.5& 96.0& 94.6& 60.5& 88.9  \\
            \cmidrule{1-1} \cmidrule{3-15}
            DAPrompt \cite{ge2022domain} & & \underline{99.2}& 92.5& 93.3& 75.4& 98.6& 92.8& 95.2& 82.5& 89.3& \underline{96.5}& 95.1& 63.5& 89.5 \\ 

            AD-CLIP \cite{singha2023ad} & & \multirowcell{-6}[-4.8ex]{\hspace*{-2.8em} \vspace{1.5em}\turnbox{90}{\thead{ViT-B/16}}} \hspace{-0.36cm}  {\bf 99.6}& \underline{92.8}& {\bf 94.0}& 78.6& \underline{98.8}& 95.4& \underline{96.8}& 83.9& 91.5& 95.8& \underline{95.5}& \underline{65.7}& \underline{90.7} \\ 

            \rowcolor{mygray} DAMP (Ours) & & 98.7& \underline{92.8}& 91.7& 80.1& {\bf 98.9}& 96.9& 94.9& 83.2& 93.9& 94.9& 94.8& {\bf 70.2}& {\bf 90.9} \\ \bottomrule
        \end{tabular}}
        \vspace{-13pt}
    \end{table*}

    \begin{table*}[t]
        \centering
        \caption{Classification accuracies (\%) on {\bf Mini-DomainNet} dataset for UDA. Marks and symbols share the same meaning in Table \ref{tab:officehome}.}
        \vspace{-5pt}
        \resizebox{0.95\linewidth}{!}{
            \begin{tabular}{l|c|cccccccccccccc}
                \toprule
                    
                   {\textbf{Method}} & $f_v$ & {Cl$\rightarrow$Pn} & {Cl$\rightarrow$Rl} & {Cl$\rightarrow$Sk} & {Pn$\rightarrow$Cl} & {Pn$\rightarrow$Rl} & {Pn$\rightarrow$Sk} & {Rl$\rightarrow$Cl} & {Rl$\rightarrow$Pn} & {Rl$\rightarrow$Sk} & {Sk$\rightarrow$Cl} & {Sk$\rightarrow$Pn} & {Sk$\rightarrow$Rl} & {\textbf{Avg}} \\ 
                    \midrule
                     ResNet-50 \cite{he2016deep} & & 52.1 & 63.0 & 49.4 & 55.9 & 73.0 & 51.1 & 56.8 & 61.0 & 50.0 & 54.0 & 48.9 & 60.3 & 56.3 \\
                     \cmidrule{1-1} \cmidrule{3-15}
                     CLIP \cite{radford2021learning}& & 67.9& 84.8& 62.9& 69.1& 84.8& 62.9& 69.2& 67.9& 62.9& 69.1& 67.9& 84.8& $71.2$ \\
                     \cmidrule{1-1} \cmidrule{3-15}
                     DAPrompt \cite{ge2022domain}& & \underline{72.4}& 87.6& 65.9& 72.7& \underline{87.6}& \underline{65.6}& 73.2& 72.4& 66.2& \underline{73.8}& 72.9& 87.8& $74.8$ \\

                     AD-CLIP \cite{singha2023ad} & & \multirowcell{-3}[-0.0ex]{\hspace*{-2.1em} \vspace{-2.8em}\turnbox{90}{\thead{ResNet-50}}} \hspace{-1.0cm} 71.7&  \underline{88.1}& \underline{66.0}& \underline{73.2}& 86.9& 65.2& \underline{73.6}& \underline{73.0}& \underline{68.4}& 72.3& \underline{74.2}& {\bf 89.3}& \underline{75.2} \\ 

                     \rowcolor{mygray} DAMP (Ours) & & {\bf 76.7} & {\bf 88.5} & {\bf 71.7} & {\bf 74.2} &	{\bf 88.7} & {\bf 70.8} & {\bf 74.4} & {\bf 75.7}&	{\bf 70.5} & {\bf 74.9}& {\bf 76.1}&	\underline{88.2}&	{\bf 77.5}
                     \\ 

            \midrule    
            \midrule
                    ViT-B \cite{dosovitskiy2020image} & & 63.3&79.0&	56.4&	62.6&	83.3&	55.4&	62.0&	70.3&	53.5&	63.0&	63.6&	75.8&	65.7      \\
                    \cmidrule{1-1} \cmidrule{3-15}
                    CLIP \cite{radford2021learning} & & 80.3& 90.5& 77.8& 82.7& 90.5& 77.8& 82.7& 80.3& 77.8& 82.7& 80.3& 90.5& $82.8$  \\
                    \cmidrule{1-1} \cmidrule{3-15}
                    DAPrompt \cite{ge2022domain} & & 83.3& 92.4& 81.1& 86.4& 92.1& 81.0& 86.7& 83.3& 80.8& 86.8& 83.5& 91.9& $85.8$ \\
                
                    AD-CLIP \cite{singha2023ad} & & \multirowcell{-3}[-0.0ex]{\hspace*{-2.1em} \vspace{-2.8em}\turnbox{90}{\thead{ViT-B/16}}} \hspace{-1.0cm} \underline{84.3}& {\bf 93.7}& \underline{82.4}& {\bf 87.5}& {\bf 93.5}& \underline{82.4}& {\bf 87.3}& \underline{84.5}& \underline{81.6}& {\bf 87.9}& \underline{84.8}& \underline{93.0}& \underline{86.9} \\   
                    \rowcolor{mygray} DAMP (Ours) & & {\bf 86.4} & \underline{93.3} &	{\bf 83.5} &	\underline{87.2} &	\underline{93.4} &	{\bf 84.1} &\underline{87.2} &	{\bf 86.5} &	{\bf 82.5} &	\underline{87.3} &	{\bf 86.6} &	{\bf 93.4} &	{\bf 87.6} \\ 
            \bottomrule
                    
                \end{tabular}}
                \vspace{-15pt}
        \label{tab:mini_domainnet}
    \end{table*}

In this section, we mainly verify the effectiveness of our method on UDA tasks. More evaluation on Multi-Source UDA \cite{peng2019moment} and Domain Generalization (DG) \cite{li2017deeper} tasks and more analytical experiments can be found in Appendix.

\subsection{Experimental Setup} 

\textbf{Datasets.} We evaluate our method on three widely used UDA datasets. {\bf Office-Home} \cite{venkateswara2017deep} consists of images from 4 different domains: Art (Ar), Clipart (Cl), Product (Pr) and Real-World (Rw). There are 65 object categories and around 15,500 images in total. {\bf VisDA-17} \cite{peng2017visda} contains synthetic images to real images across 12 categories. The synthetic source domain has 152,397 images generated from 3D models. The real target domain has 55,388 real images. {\bf Mini-DomainNet} is a subset of the most challenging dataset DomainNet \cite{peng2019moment}. We use a subset with 4 domains, i.e., Clipart (Cl), Painting (Pn), Real (Rl) and Sketch (Sk), across 126 categories following previous works \cite{zhou2021domain,singha2023ad}.

\textbf{Training Configuration.} We evaluate DAMP with both ResNet-50 \cite{he2016deep} and ViT-B/16 \cite{dosovitskiy2020image} as the visual encoder $f_v$. The text encoder $f_s$ is a pretrained CLIP text encoder with depth $J=12$. During training, we freeze these encoders and tune the input textual prompts $\bm{p}_{1:N}$ and the prompting module $G$. The learnable token length $N$ is set to 32 and we use $L=2$ Transformer decoder layers in the mutual prompting module $G$. For training, we use the Adam optimizer \cite{kingma2014adam} with an initial learning rate of 3e-3 for all datasets, and adjust it with a cosine annealing scheduler \cite{loshchilov2016sgdr}. Our model is trained for 30 epochs in total (for Mini-DomainNet, we train 500 iterations per epoch). The batch size is set to 32 for each domain. We set the confidence threshold $T=0.6$ for Office-Home and $0.5$ for VisDA-17 and Mini-DomainNet. For hyperparameters, we use $\lambda_c = \lambda_i = 1.0$. Due to the modality gap in CLIP \cite{jiang2023understanding,liang2022mind}, the visual and textual embeddings need different updating magnitudes, making manually searching $\gamma_v$ and $\gamma_s$ challenging. Therefore, we make them learnable parameters. More implementation details about network architectures and pseudo-labels can be found in Appendix.

\subsection{Comparasion with State-of-the-Arts} 
We report the results on {\bf Office-Home} in Table \ref{tab:officehome}. Our DAMP outperforms all competitors on most tasks, especially the challenging ones like Ar$\rightarrow$Cl and Cl$\rightarrow$Ar. Besides, DAMP brings substantial gains over strong baselines, improving the average accuracy over PADCLIP by 1.6\% with ResNet-50 and 0.4\% with ViT-B. Compared to prompt-based methods like DAPrompt and AD-CLIP, DAMP also shows superiority by mutually aligning both modalities. For example, it surpasses DAPrompt by 3.7\% with ResNet-50 and 3.1\% with ViT-B. The improvements are more significant with ResNet-50. This indicates DAMP can better exploit ViT's intrinsic transferability while effectively prompting ResNet for inspiring improvements.

For {\bf VisDA-17} (Table \ref{tab:visda}), DAMP demonstrates competitive performance compared to other methods. It achieves 88.4\% average accuracy with ResNet-101. Although it is slightly worse than the best competitor PADCLIP (88.5\%), we show in Sec. \ref{sec:analysis} that our method involves much less learnable parameters. When using ViT-B, DAMP obtains the highest average accuracy (90.9\%), comparable to PADCLIP. The results validate the consistently strong performance of DAMP across different vision backbones.

As shown in Table \ref{tab:mini_domainnet}, DAMP sets new state-of-the-art on {\bf Mini-DomainNet}. It brings significant improvements over baseline methods like standard CLIP and other prompt learning methods like DAPrompt and AD-CLIP. For example, with ResNet-50, DAMP improves over CLIP by 6.3\%, DAPrompt by 2.7\%, and AD-CLIP by 2.3\%. Similar gains are observed when using ViT backbone. In addition, it achieves especially large gains on challenging tasks like Cl$\rightarrow$Pn, Cl$\rightarrow$Sk and Sk$\rightarrow$Pn. The consistent and substantial improvements of DAMP over strong baselines highlight the benefits of mutually aligning textual and visual prompts in a domain-agnostic and instance-conditioned manner.

\begin{figure}[t]  
    \centering 
    \subfloat[Visual embeddings]{
    \includegraphics[width=0.45\linewidth] {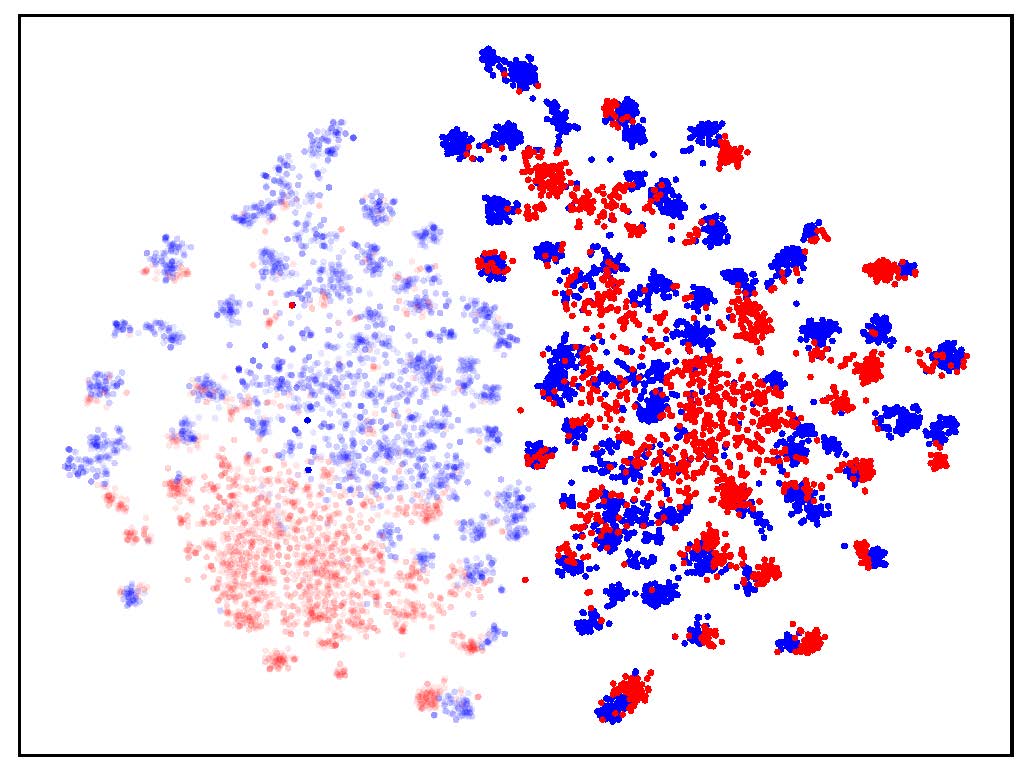}
    \label{fig:tsne_visual}
    }
    \subfloat[Textual embeddings]{
    \includegraphics[width=0.45\linewidth] {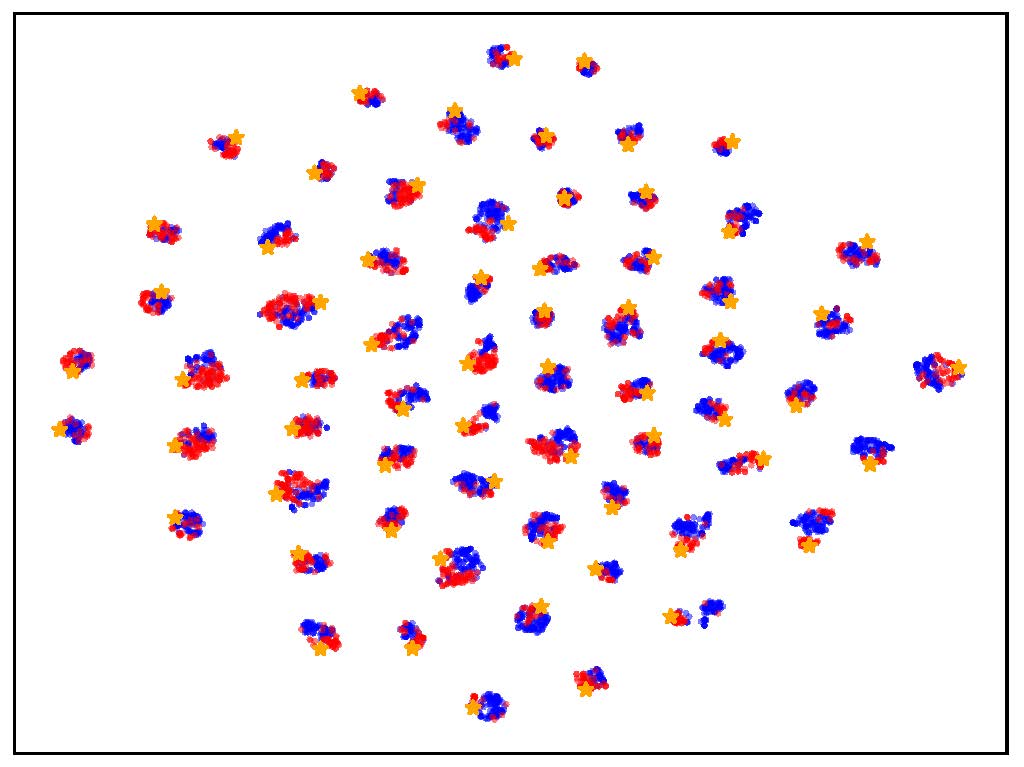}
    \label{fig:tsne_textual} 
    } 
    \vspace{-4pt}  
    \caption{Visualization of (a) visual embeddings and (b) textual embeddings using t-SNE \cite{van2008visualizing} on task Ar $\rightarrow$ Pr (Office-Home). Light and dark colors represent embeddings before and after our mutual prompting, respectively. Red and blue points are source and target samples, respectively. Orange stars denote the class-level domain-agnostic textual embeddings $\{\bm{s}_k\}_{k=1}^{K}$.}
    \label{fig:tSNE_visualization}
    \vspace{-15pt}
  \end{figure}

\subsection{Analytical Experiments} \label{sec:analysis}
\textbf{Visualization of Embeddings.} Fig. \ref{fig:tSNE_visualization} visualizes the visual and textual embeddings learned by DAMP. In Fig. \ref{fig:tsne_visual}, we see that before visual prompting, the source and target domains form distinct distributions, indicating a large domain gap. After prompting, the visual features become better aligned across domains and form more clear clusters, suggesting a reduced domain gap and a discriminative semantic structure. Fig. \ref{fig:tsne_textual} shows that instance-conditioned textual prompting increases within-class semantic diversity. This enables better pairing of text and images in both domains. Through mutual alignment of visual and textual embeddings, the two prompts make representations more domain-invariant to facilitate cross-domain knowledge transfer.

\begin{figure}[t]
    \centering
     \includegraphics[width=0.6\linewidth]{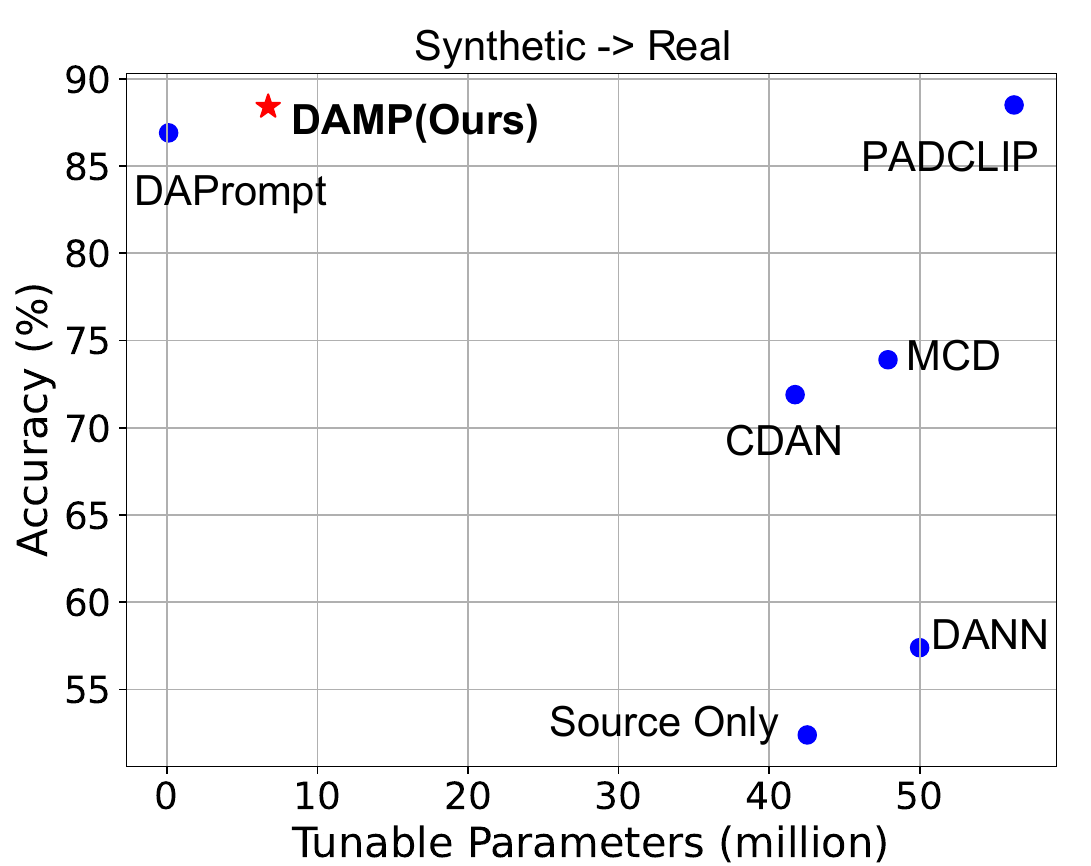}
     \vspace{-4pt}
     \caption{Comparasion between different UDA methods regarding tunable parameters and accuracies on VisDA-17 (ResNet-101). DAMP only use 11.9\% parameters compared with PADCLIP.}
     \label{fig:parameter_acc}
  \vspace{-8pt}
  \end{figure}

\textbf{Model Capacity Analysis.} Fig. \ref{fig:parameter_acc} compares different UDA methods regarding the number of tunable parameters versus accuracy. Traditional methods like DANN and CDAN fine-tune the full model, requiring extensive parameters yet achieving relatively low accuracy. PADCLIP \cite{lai2023padclip} fine-tunes the CLIP visual backbone, introducing many parameters. In contrast, our DAMP only tunes the textual prompts $\bm{p}_{1:N}$ and the prompting module $G$, which just adds a few parameters over DAPrompt yet achieves comparable accuracy (88.4\%) to PADCLIP.

  \begin{table}[t]
    \vspace{-2pt} 
    \centering
    \scriptsize
    \caption{Comparasion between different prompting strategies on Office-Home (ViT-B/16). MP denotes mutual prompting.}
    \vspace{-3pt} 
    \setlength{\tabcolsep}{5mm}{
        \begin{tabular}{lc}
            \toprule
            Prompting Strategy                   & Office-Home \\ \midrule
            w/o MP (CoOp)            & 85.0        \\ 
            Independent prompting    & 85.4        \\ 
            MP w/ simple synergy  & 85.9        \\
            MP w/ cross-attention & {\bf 86.1}        \\ \bottomrule
            \end{tabular}}
    \label{tab:prompting_strategy}
    \vspace{-15pt}
  \end{table}

  \begin{figure*}[t] 
    \centering 
    \subfloat[$N$ (ResNet-101)]{
    \includegraphics[width=0.228\linewidth] {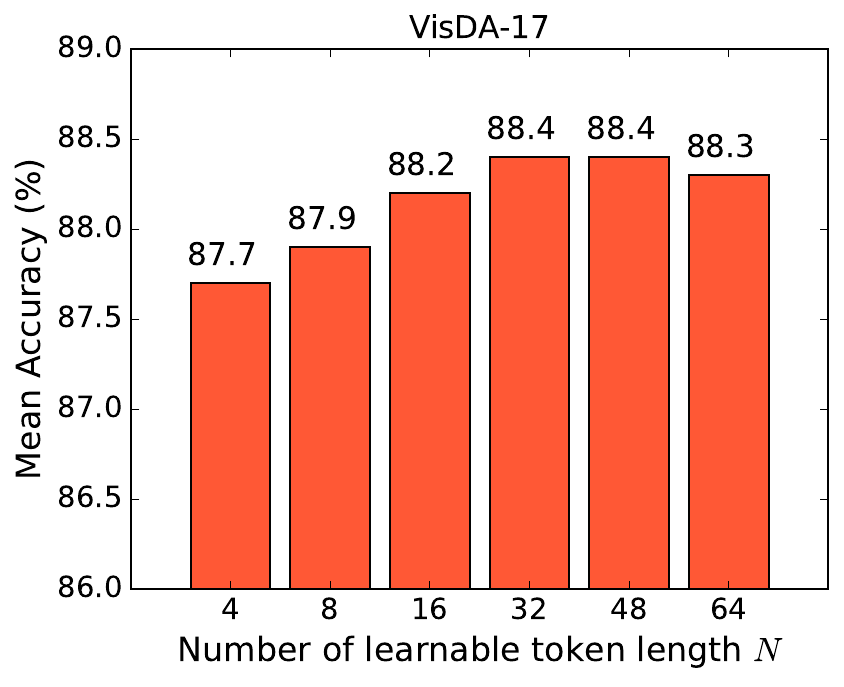} 
    \label{fig:img_N_UDA}
    }
    \subfloat[$\gamma_v$ and $\gamma_t$ (ResNet-50)]{
    \includegraphics[width=0.225\linewidth] {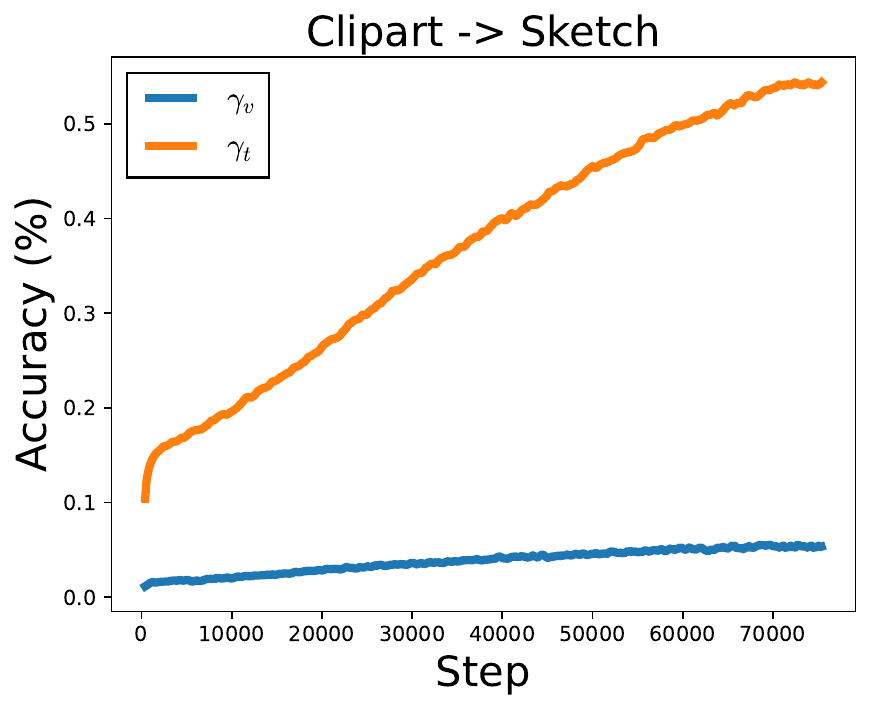}
    \label{fig:img_gamma}  
    }
    \subfloat[$T$ (ResNet-50)]{
    \includegraphics[width=0.223\linewidth] {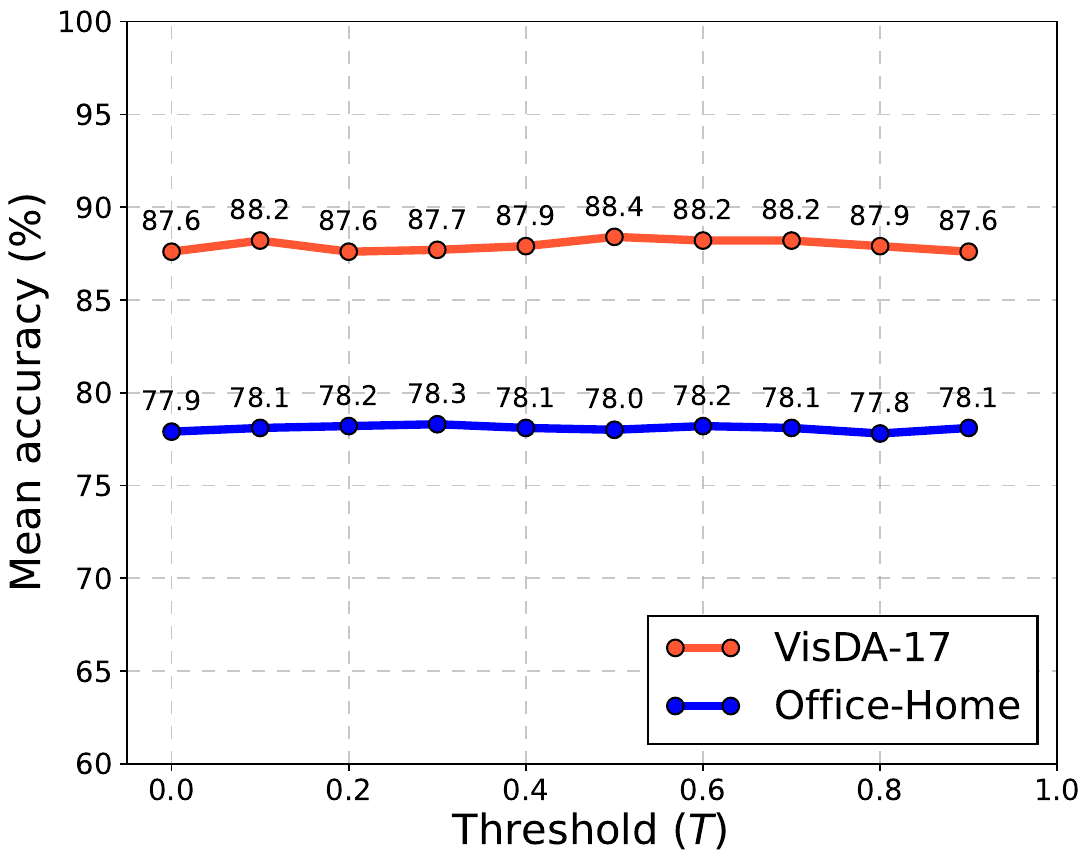}
    \label{fig:img_threshold}
    }
    \subfloat[$\lambda_c$ and $\lambda_i$ (ResNet-50)]{
    \includegraphics[width=0.22\linewidth] {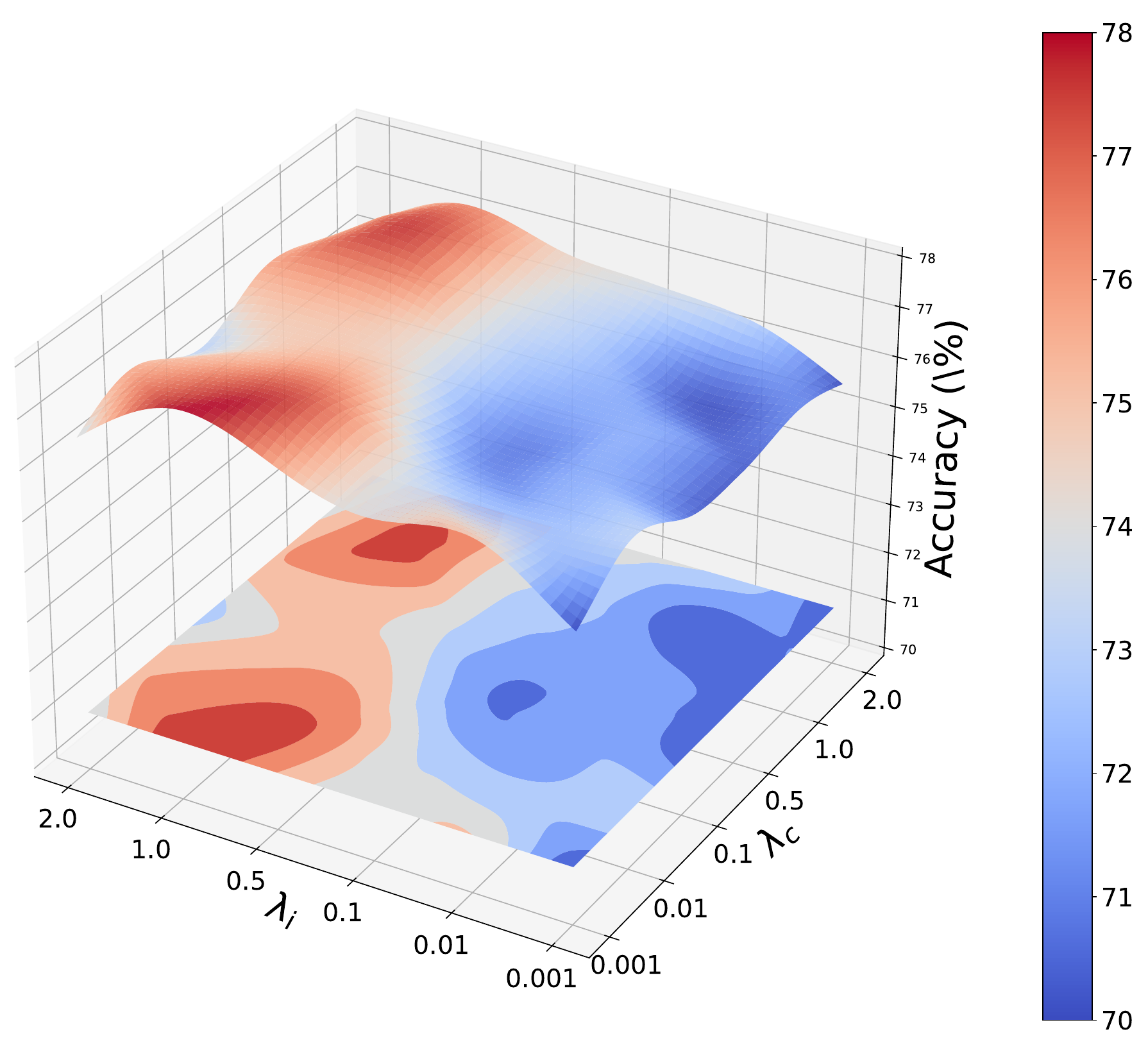}
    \label{fig:img_para_sen}
    }
    \vspace{-5pt}
    \caption{Hyperparameter analysis. (a) Performance under different learnable token length $N$ on VisDA-17 dataset. (b) Values of learnable hyperparameters $\gamma_v$ and $\gamma_t$ during training on task Cl $\rightarrow$ Sk (Mini-DomainNet). (c) The influence of different choices of $T$ on Office-Home dataset. (d) parameter sensitivities of $\lambda_c$ and $\lambda_i$ on task Cl $\rightarrow$ Ar (Office-Home).} 
    \label{fig:parameter_analysis} 
    \vspace{-14pt} 
  \end{figure*}

  \textbf{Analysis of the Mutual Prompting Strategy.} To further explore the effectiveness of the cross-attention-based mutual prompting framework, we conduct experiments comparing it to other prompting strategies in Table \ref{tab:prompting_strategy}. For fair comparison, all regularizations are removed in this experiment and only $\mathcal{L}_{sup}$ is optimized. Specifically, we evaluate an independent prompting strategy by replacing the cross-attention with self-attention in $G$, which results in separate prompting modules for each modality without any interaction. It shows that without mutual synergy, the improvement is limited over the baseline CoOp. We also examine a simple synergy strategy inspired by MaPLe \cite{khattak2023maple}, where we use a linear projection layer to obtain the prompted textual embedding $\bm{s}^*_k$ from the image context embedding $\bm{\widetilde{v}}$ (and vice versa for the visual embedding). However, this uni-directional projection does not fully capture the complex interaction between the modalities in the prompting process. In contrast, the cross-attention module in our framework allows bi-directional interaction, which enables more effective fusion of information from both modalities, resulting in better domain-agnostic and instance-specific prompts.

  \begin{table}[t]
    \vspace{-2pt}
    \centering
    \scriptsize
    \caption{Ablation study on Office-Home (ResNet-50) and VisDA-17 (ResNet-101). ITP and VP refer to instance-level textual prompting and visual prompting, respectively.}
    \vspace{-3pt}
    \resizebox{0.42\textwidth}{!}{
    \setlength{\tabcolsep}{0.55mm}{
        \begin{tabular}{l|ccccc|cc}
            \toprule
            \multirow{2}{*}{Method}           & \multicolumn{2}{c|}{Prompting}        & \multicolumn{3}{c|}{Loss} & \multicolumn{2}{c}{Mean Accuracy} \\ \cmidrule{2-8} 
                                              & ITP & \multicolumn{1}{c|}{VP} & $\mathcal{L}_{sc}$  & $\mathcal{L}_{idc}$   & $\mathcal{L}_{im}$    & Office-Home       & VisDA-17      \\ \midrule
            Baseline (CoOp)                   & \ding{55}       & \ding{55}                           & \ding{55}      & \ding{55}       & \ding{55}      & 75.3              & 86.1          \\ \midrule
            \multirow{2}{*}{Uni-modal Prompting} & \ding{55}       & \checkmark                           & \ding{55}      & \ding{55}       & \ding{55}      & 75.7              & 86.5          \\ \cmidrule{2-8} 
                                              & \checkmark       & \ding{55}                           & \ding{55}      & \ding{55}       & \ding{55}      & 75.8              & 87.1          \\ \midrule
            \multirow{4}{*}{Mutual Prompting}             & \checkmark       & \checkmark                           & \ding{55}      & \ding{55}       & \ding{55}      & 76.1              & 87.2          \\ \cmidrule{2-8} 
                                              & \checkmark       & \checkmark                           & \checkmark      & \ding{55}       & \ding{55}      & 76.9              & 87.8          \\ \cmidrule{2-8} 
                                              & \checkmark       & \checkmark                           & \checkmark      & \checkmark       & \ding{55}      & 77.3              & 88.0          \\ \cmidrule{2-8} 
                                              & \checkmark       & \checkmark                           & \checkmark      & \checkmark       & \checkmark      & 78.2              & 88.4          \\ \bottomrule
              \end{tabular}}}
    \label{tab:ablation} 
    \vspace{-15pt}
  \end{table}

  \textbf{Hyperparameter Analysis.} There are four categories of hyperparameters in our method, i.e., number of learnable tokens $N$, learnable weights $\gamma_v$ and $\gamma_s$, confidence threshold $T$, and loss weights $\lambda_c$ and $\lambda_i$. As shown in Fig. \ref{fig:img_N_UDA}, the performance on VisDA-17 improves as $N$ increases from 4 to 32, and then remains relatively stable with larger $N$. This indicates that a moderate length is enough for encoding rich semantic information. However, further increasing $N$ does not bring substantial gains.

  Fig. \ref{fig:img_gamma} plots the values of the learnable weight coefficients $\gamma_v$ and $\gamma_s$ during training. We can see that $\gamma_v$ converges to a relatively small value around 0.1, while $\gamma_s$ converges to a larger value around 0.5. This aligns with the intuition that a smaller perturbation is needed on the visual embeddings compared to the textual embeddings, due to the inherent modality gap between vision and language in CLIP. The learnable nature of $\gamma_v$ and $\gamma_s$ provides the flexibility to adapt the updating magnitudes for both modalities.
  
  Fig. \ref{fig:img_threshold} shows that the performance is relatively stable across different choices of $T$ from 0.5 to 1.0. Setting $T$ too small (0.1) deteriorates the performance. This indicates that the confidence threshold is not too sensitive, while using very unconfident pseudo-labels can hurt the performance. A moderate threshold between 0.5 and 1.0 works reliably.
  
  Fig. \ref{fig:img_para_sen} studies $\lambda_c$ and $\lambda_i$ on task Cl$\rightarrow$Ar. The results show DAMP is relatively robust to different choices of $\lambda_c$, and a little sensitive to the variation of $\lambda_c$. Overally, trivially setting both of them to 1.0 offers a good trade-off.

  \textbf{Ablation Study.} To validate the efficacy of each component in our method, we conduct an ablation study on Office-Home and VisDA-17 datasets. As shown in Table \ref{tab:ablation}, we first evaluate a baseline model that directly optimizes the supervised loss $\mathcal{L}_{sup}$ on source and confident target samples to fine-tune $\bm{p}_{1:N}$ without further prompting (analogous to CoOp), which gives the lowest performance. Adding either instance-level textual prompting (ITP) or visual prompting (VP) brings gains over the baseline, showing the benefits of adapting either modality with prompting. Further prompting both modalities together with the mutual prompting framework leads to additional performance boosts, which mainly benefits from the flexibility to adapt both language and vision branches to the target domain. Finally, regularizations $\mathcal{L}_{sc}$, $\mathcal{L}_{idc}$ and $\mathcal{L}_{im}$ all contribute to the superior performance in a collaborative manner. The step-wise improvements support the rationality of our design.
\section{Conclusion}
We propose DAMP, a novel framework to address UDA using VLMs. DAMP mutually aligns the visual and textual modalities via prompting to elicit domain-agnostic embeddings. The prompts are optimized together through cross-attention and regularized with elaborate losses. Extensive experiments validate that DAMP brings substantial and consistent improvements over strong baselines on three benchmarks. DAMP provides an effective approach to harness both source and pre-trained VLMs knowledge for UDA.

{
    \small
    \bibliographystyle{ieeenat_fullname}
    \bibliography{main}
}

\clearpage
\setcounter{page}{1}
\maketitlesupplementary
\appendix

\section*{Appendix Contents}

\begin{itemize}
    \item[A.] \nameref{sec:appendix_Implementation}
    \item[B.] \nameref{sec:appendix_Algorithm}
    \item[C.] \nameref{sec:appendix_msda}
    \item[D.] \nameref{sec:appendix_dg}
    \item[E.] \nameref{sec:appendix_analytical}
\end{itemize}

\section{More Implementation Details}
\label{sec:appendix_Implementation}
We implement our method with the Pytorch framework. Our code is built on the \texttt{Dassl.pytorch} \footnote{https://github.com/KaiyangZhou/Dassl.pytorch} platForm, which is a principled implementation and evaluation platForm For DA and DG tasks. We train DAMP with a single NVIDIA GeForce RTX 3090 GPU. Details about network architecture, data augmentation and pseudo-labels are described as follows.

\textbf{Network Architecture.} The text encoder $f_s$ and the mutual prompting module $G$ we used are mainly comprised of a TransFormer encoder and a TransFormer decoder, respectively. Specifically, the standard dot-product attention is leveraged. Given a set of queries $Q \in \mathbb{R}^{N_q \times d_k}$, keys $K \in \mathbb{R}^{N_k \times d_k}$ and values $V \in \mathbb{R}^{N_k \times d_v}$, the attentional outputs For all queries can be calculated by:
\begin{equation}
    \texttt{Attention}(Q, K, V)=\operatorname{softmax}\left(\frac{Q K^T}{\sqrt{d_k}}\right) V \in \mathbb{R}^{N_q \times d_v}.
    \end{equation}
For Self-Attention (SA), $Q, K, V$ obtained from the same input sequence $I \in \mathbb{R}^{N \times D}$, through three projection matrixes $M_q \in \mathbb{R}^{D \times d_k}, M_k \in \mathbb{R}^{D \times d_k}, M_v \in \mathbb{R}^{D \times d_v}$,
\begin{equation}
    \texttt{SA}(I)=\texttt{Attention}(I M_q, I M_k, I M_v) \in \mathbb{R}^{N \times d_v}.
    \end{equation}
Multi-Head Self-Attention (MHSA) extends SA by using multiple attention heads,
\begin{equation}
    \texttt{MHSA}(I) =\operatorname{Concat}\left(\operatorname{head}_1, \ldots, \operatorname{head}_{\mathrm{h}}\right) M_O, \\
    \end{equation}
where $\operatorname{head}_i =\texttt{Attention}\left(I M^i_q, I M^i_k, I M^i_v\right)$, $h$ is the head number, and $M_O \in \mathbb{R}^{h d_v \times D}$ maps the intermediate embeddings to match the input dimension.

For $f_s$, each encoder layer $\texttt{Enc}_j$ comprised of a MHSA block and a feed-Forward block $\texttt{FD}(\cdot)$, with $h=8$, $d_k=d_v=D=512$. A residual connection and LayerNorm operation $\texttt{LN}(\cdot)$ is employed after each of them, i.e.,
\begin{equation}
    \begin{aligned}
        & \texttt{Enc}_j (I) = \texttt{LN}(\texttt{FD}(I^{\prime}) + I^{\prime}), \\
        & I^{\prime} = \texttt{LN}(\texttt{MHSA}(I) + I)
    \end{aligned}
\end{equation}

For $G$, a Masked MHSA (M-MHSA) block and Multi-Head Cross-Attention (MHCA) block is used in each $\texttt{Dec}_l$. Specifically, the M-MHSA alters MHSA by imposing a mask on the attention scores, i.e.,
\begin{equation}
    \begin{aligned}
    &\texttt{M-MHSA}(I) =\operatorname{Concat}\left(\operatorname{head}_1, \ldots, \operatorname{head}_{\mathrm{h}}\right) M_O, \\
    &\operatorname{head}_i =\texttt{MaskedAttention}\left(I M^i_q, I M^i_k, I M^i_v\right), \\
    &\texttt{MaskedAttention}(Q, K, V)=\operatorname{softmax}\left(\frac{S \odot Q K^T}{\sqrt{d_k}}\right), \\
\end{aligned}
\end{equation}
where $S \in \mathbb{R}^{N_q \times N_k}$ is the mask. The MHCA block is a variation of MHSA by using different source of $Q$ and $K,V$, i.e., 
\begin{equation}
\begin{aligned}
   & \texttt{MHCA}(I_1,I_2) =\operatorname{Concat}\left(\operatorname{head}_1, \ldots, \operatorname{head}_{\mathrm{h}}\right) M_O, \\
   & \operatorname{head}_i =\texttt{Attention}\left(I_1 M^i_q, I_2 M^i_k, I_2 M^i_v\right).
\end{aligned}    
\end{equation}
Each decoder layer $\texttt{Dec}_l$ can be Formulated as,
\begin{equation}
    \begin{aligned}
        & \texttt{Dec}_l (I_1, I_2) = \texttt{LN}(\texttt{FD}(I_1^{\prime\prime}) + I_1^{\prime\prime}), \\
        & I_1^{\prime\prime} = \texttt{LN}(\texttt{MHCA}(I_1, I_2) + I_1), \\
        & I_1^{\prime} = \texttt{LN}(\texttt{M-MHSA}(I_1) + I_1).
    \end{aligned}    
\end{equation}
We use $d_k=d_v=D=256$ and $h=4$ For the decoder. The $\texttt{InProj}$ and $\texttt{OutProj}$ are two linear layers with LayerNorm operations. In this work, we use the same $G$ For language-guided visual prompting and vision-guided language prompting since the two modalities are aligned in the CLIP embedding space. Therefore, they can attend to each other by freely change the position in $\texttt{Dec}_l$.

\textbf{Data augmentations.} In this work, we use random flip as a simple weak augmentation operation. For strong augmentation, we select two random operations For each sample from the RandAugment library \cite{cubuk2020randaugment}, which includes invert, rotation, color enhancing, auto contrast and other transFormations. Randomly selecting and combining these strong augmentations is intended to simulate the diverse domain shifts that can occur in real-world data.

\textbf{Pseudo-label.} For previous methods that incorporate domain-specific tokens in the textual prompt, obtaining pseudo-labels for the target domain is challenging. On one hand, the domain-specific prompts are not transferable, i.e., we cannot obtain high-quality target pseudo-labels with the learned source-specific prompts. On the other hand, without high-quality pseudo-labels, learning target-specific prompts becomes an ill-posed problem. To circumvent this problem, DAPrompt only \cite{ge2022domain} uses a naive textual prompt, i.e., ``a photo of a [CLS]. a [Domain] image.'' to obtain pseudo-labels, where [CLS] and [Domain] are the name of each class and each domain, respectively. This resuls in knowledge isolation between the two domains. 

In this work, we learn shared prompts For both domains. This allows us to leverage the rich source domain knowledge to pseudo-label the target domain. However, we found that in the early stages of training, the source domain model is not yet well-trained, resulting in low-quality pseudo-labels. To address this, we propose combining prior knowledge with the source knowledge to obtain better pseudo-labels. Specifically, we first generate a naive textual prompt to produce naive soft pseudo-labels $\tilde{y}^t_i$ for each target sample $y^t_i$. We also generate source-enabled soft pseudo-labels $\dot{y}^t_i$ from the model outputs according to Eq. (7). The final pseudo label is an ensemble of both:
\begin{equation}
\hat{y}^t_i = (1-\alpha) \tilde{y}^t_i + \alpha \dot{y}^t_i.
\end{equation}
The weight $\alpha$ is gradually increased from 0 to 1 during training. Weighting the naive and source-enabled pseudo-labels via the $\alpha$ enables a smooth transition from relying more on the prior knowledge to relying more on the source-knowledge as training progresses.

\begin{algorithm}[t]
    \caption{Training Procedure of DAMP For UDA.}
    \label{alg:DAMP}
    \begin{algorithmic}[1]
        \Require Labeled source dataset $\mathcal{D}_s$, unlabeled target dataset $\mathcal{D}_t$, total training epochs $E$, iteration number per epoch $N_e$.
        \Ensure Optimal $\bm{p}_{1:N}$, $\gamma_v$, $\gamma_s$ and $G$.
        \State Initialize parameters for $\bm{p}_{1:N}$, $\gamma_v$, $\gamma_s$ and $G$.
        \For{$t=1$ {\bfseries to} $E$}
        \For{$i=1$ {\bfseries to} $N_e$}
            \State Sample a source batch $\mathcal{B}_s \sim \mathcal{D}_s$ and a target batch $\mathcal{B}_t \sim \mathcal{D}_t$
            \State obtain $\{\bm{s}_k^{\prime}\}_{k=1}^K$ and $\bm{v}^{\prime}$ for each $\bm{x} \in \mathcal{B}_s \cup \mathcal{B}_t$ according to Eq. (6) and (4).
            \State Calculate $\mathcal{L}^s_{sup}$ and $\mathcal{L}^t_{sup}$ according to Eq. (12) and (13).
            \State Calculate $\mathcal{L}^s_{sc}$ and $\mathcal{L}^t_{sc}$ according to Eq. (9) and (10).
            \State Calculate $\mathcal{L}_{idc}$ within $\mathcal{B}_s$ and $\mathcal{B}_t$ respectively according to Eq. (8) and sum them up.
            \State Calculate $\mathcal{L}_{im}$ according to Eq. (11).
            \State Update parameters via optimizing $\mathcal{L}_{all}$. 
        \EndFor
        \EndFor

        \State Return final model parameters $\bm{p}_{1:N}$, $\gamma_v$, $\gamma_s$ and $G$.
    \end{algorithmic}
\end{algorithm}

\section{Algorithm}
\label{sec:appendix_Algorithm}
To better understand our method, we summarize the training procedure of DAMP for UDA in Algorithm \ref{alg:DAMP}.

\section{Experiments on Multi-Source UDA} 
\label{sec:appendix_msda}
\begin{table*}[t]
    \begin{center}
    \centering
    \caption{Classification accuracies (\%) on {\bf DomainNet} for MSDA with ResNet-101. * Prompt learning-based methods.}
    \footnotesize
    \setlength{\tabcolsep}{4.5mm}{
    \begin{tabular}{lccccccc}
    \toprule
     {\bf Method} & $\rightarrow$Clipart & $\rightarrow$Infograph  & $\rightarrow$Painting & $\rightarrow$Quickdraw & $\rightarrow$Real  & $\rightarrow$Sketch  &Avg. \\
      \midrule
    
      {\bf Zero-Shot} &   &  &  &  &  &  &  \\
      ~~CLIP \cite{radford2021learning} & 61.3  &42.0  &56.1  &10.3  &79.3  &54.1  &50.5  \\
      \midrule \midrule
      {\bf Source Combined} &   &  &  &  &  &  &  \\
    ~~DANN \cite{ganin2015unsupervised}  &45.5     &13.1     &37.0     &13.2    &48.9    &31.8    &32.6\\
    ~~MCD \cite{saito2018maximum}  &54.3     &22.1     &45.7     &7.6    &58.4    &43.5    &38.5\\
    ~~DAPrompt * \cite{ge2022domain} & 62.4  &43.8  &59.3   &10.6   &  81.5 &54.6  &52.0\\  
    ~~CoOp * \cite{zhou2022learning} & 63.1  & 41.2  & 57.7   & 10.0   & 75.8 &55.8  & 50.6 \\ 
      \midrule \midrule
      {\bf Multi-Source} &   &  &  &  &  &  &  \\
    
    ~~$\text{M}^3$SDA-$\beta$ \cite{peng2019moment}   &58.6     &26.0     &52.3     &6.3    &62.7    &49.5    &42.6\\
    ~~SImpA$\text{I}_{101}$ \cite{venkat2020your}  &66.4     &26.5     &56.6     &18.9   &68.0    &55.5    &48.6\\
    ~~LtC-MSDA \cite{wang2020learning}  &63.1     &28.7     &56.1     &16.3    &66.1    &53.8    &47.4\\
    ~~T-SVDNet \cite{li2021t}  &66.1     &25.0     &54.3     &16.5    &65.4    &54.6    &47.0\\
    ~~PFSA \cite{fu2021partial}  &64.5     &29.2     &57.6     &\textbf{17.2}    &67.2    &55.1    &48.5\\
    ~~PTMDA \cite{ren2022multi}  &66.0     &28.5     &58.4     &13.0    &63.0    &54.1    &47.2\\
    ~~MPA * \cite{chen2022multi}   &65.2     &47.3     & 62.0     &10.2   & 82.0    &57.9    & 54.1\\
    \rowcolor{mygray} ~~DAMP * (Ours)   & {\bf 69.7}     &   {\bf 51.0}   &\textbf{67.5}     & 14.7  &\textbf{82.5}    & {\bf 61.5}  &\textbf{57.8}\\
    \bottomrule
    \end{tabular}}
    \label{tab:domainnet_msda}
    \end{center}
    \vspace{-3ex}
    \end{table*}

    \begin{table}[t]
        \centering
        \footnotesize
        \caption{Classification accuracies (\%) on {\bf Office-Home} for MSDA with ResNet-50. * Prompt learning-based methods.}
        \setlength{\tabcolsep}{2.8mm}{
        \begin{tabular}{lccccc}
        \toprule
         {\bf Method} &  $\rightarrow$Ar & $\rightarrow$Cl  & $\rightarrow$Pr & $\rightarrow$Rw & Avg. \\
         \midrule
         \multicolumn{1}{l}{\textbf{Zero-Shot}} \\
        ~~CLIP \cite{radford2021learning} &71.5  &50.2 &81.3   &82.4  &71.4\\
         \midrule
         \multicolumn{1}{l}{\textbf{Source Combined}} \\
        ~~DAN \cite{long2015learning}    & 68.5     &59.4     &79.0     &82.5   &72.4\\
        ~~DANN \cite{ganin2015unsupervised}   & 68.4     &59.1     &79.5     &82.7    &72.4\\
        ~~CORAL \cite{sun2016deep}  & 68.1     &58.6     &79.5     &82.7   &72.2\\
        ~~DAPrompt * \cite{ge2022domain}  &  72.8  &51.9   &82.6  &83.7  &72.8\\
        ~~CoOp * \cite{zhou2022learning} &  70.7  &52.9   &82.9  &83.9  &72.4\\
         \midrule
         \multicolumn{1}{l}{\textbf{Multi-Source}} \\
        ~~MDDA \cite{zhao2020multi}    & 66.7     &$\textbf{62.3}$     &79.5     &79.6   &71.0 \\
        ~~SImpA$\text{I}_{50}$ \cite{venkat2020your} & 70.8 & 56.3 & 80.2 & 81.5 & 72.2\\
        ~~MFSAN \cite{zhu2019aligning} & 72.1     &62.0     &80.3     &81.8   &74.1\\
        ~~MPA * \cite{chen2022multi}  & 74.8     &54.9     &86.2     &85.7   &75.4 \\
        \rowcolor{mygray} ~~DAMP * (Ours)   & {\bf 77.7}     & 61.2   &\textbf{90.1}  &\textbf{87.7}    & {\bf 79.2} \\
        \bottomrule
        \end{tabular}}
        \label{tab:officehome_msda}
        \end{table}

To evaluate the versatility of our method in various domain adaptation scenarios, we extend our method to the multi-source domain adaptation (MSDA) setting.

\textbf{Datasets.} We evaluate our method on two widely used MSDA datasets. Specifically, we reuse the {\bf Office-Home} \cite{venkateswara2017deep} dataset for MUDA by combining arbitrary 3 domains as source domains and regard the rest domain as the target domain, which forms 4 adaptation tasks ($\rightarrow$Ar, $\rightarrow$Cl, $\rightarrow$Pr, $\rightarrow$Rw). {\bf DomainNet} \cite{peng2019moment} is the lagest and the most challenging dataset for domain adaptation. It consists of 6 diverse domains, including Clipart, Painting, Real, Sketch, Quickdraw and Infograph. These domains encompass a wide range of visual styles, making the dataset challenging for domain adaptation tasks. Similar to Office-Home, 6 tasks are constructed for MSDA. 

\textbf{Experimental Setup.} It is a natural extension to apply our method to the MSDA setting, where the goal is to learn a shared set of prompts across all source domains and the target domain. Specifically, we extend $\mathcal{L}_{sup}$ and $\mathcal{L}_{sc}$ to include losses on all source domains. We treat each domain with equal importance. For $\mathcal{L}_{idc}$, we obtain a batch of samples from each domain in each iteration, and compute $\mathcal{L}_{idc}$ within each domain batch. The utilization of domain labels in this process distinguishes our method from other single-source domain adaptation methods which simply mix the source domains. For convenient comparison with previous methods, we use the ResNet-50 backbone on Office-Home and ResNet-101 on DomainNet. Other training configurations remain consistent with those employed in single-source UDA, as detailed in Sec. 4. 

\textbf{Experimental Results.} The results on {\bf DomainNet} are reported in Table \ref{tab:domainnet_msda}. DAMP achieves the best average accuracy of 57.8\%, outperforming the previous state-of-the-art MPA by 3.7\%. This demonstrates the effectiveness of DAMP on multi-source domain adaptation. Compared to single-source methods like DANN, MCD and DAPrompt, DAMP brings substantial gains, improving over DAPrompt by 5.8\%. This shows the benefits of domain-agnostic prompts and exploiting multiple source domains in our method. Besides, DAMP also surpasses other multi-source domain adaptation methods such as $\text{M}^3$SDA-$\beta$, SImpA, LtC-MSDA and T-SVDNet by a large margin. Notably, DAMP achieves the best performance on 5 out of 6 tasks. The consistent improvements over competitive baselines validate the robustness of DAMP.

On {\bf Office-Home} (Table \ref{tab:officehome_msda}), we can observe that DAMP again achieves state-of-the-art accuracy (79.2\%), outperforming the closest competitor MPA by 3.8\%. Compared to single-source methods, DAMP brings significant gains over 6.4\% over DAPrompt, showing the benefit of learning domain-agnostic prompts in the multi-source setting. DAMP surpasses other multi-source domain adaptation methods like MDDA, SImpA and MFSAN by solid margins. Furthermore, by comparing with the results for single-source UDA (Table 1), we can observe that for any target domain, using multiple source domain data is better than using only one source domain. This validates that our method indeed utilizes source domain knowledge effectively. 

\section{Experiments on Doamin Generalization} 
\label{sec:appendix_dg}

\begin{table*}[t]
    \caption{
        Classification accuracies (\%) on VLCS, PACS, Office-Home, and TerraIncognita for domain generalization. The best results are highlighted in bold. All compared methods are implemented based on CLIP with ViT-B/16 backbone.}
    \vspace{-6pt}
    \begin{center}
    \setlength{\tabcolsep}{3mm}{
    \begin{tabular}{lccccc}
    \toprule
    \textbf{Method} & VLCS & PACS & Office-Home & TerraInc & Avg \\
    \midrule
    \textbf{Fine-tuning (CLIP)} & & & & & \\
    ~~ERM & 82.7 $\pm$ 0.3 & 92.9 $\pm$ 1.9 & 78.1 $\pm$ 2.1 & 50.2 $\pm$ 1.7 & 75.9\\
    ~~CORAL \cite{sun2016deep} & 82.0 $\pm$ 0.2 & 93.2 $\pm$ 1.1 & 78.9 $\pm$ 1.9 & \textbf{53.5 $\pm$ 0.7} & 76.9 \\
    ~~DANN \cite{ganin2015unsupervised} & 83.2 $\pm$ 1.2 & 93.8 $\pm$ 1.3 & 78.8 $\pm$ 1.1 & 52.2 $\pm$ 2.0 & 77.0 \\
    \midrule
    \textbf{Zero-shot} & &  & & & \\
    ~~CLIP \cite{radford2021learning} & 82.3 $\pm$ 0.1 & 96.1 $\pm$ 0.1 & 82.3 $\pm$ 0.2 & 34.1 $\pm$ 0.1 & 73.7 \\
    \midrule
    \textbf{Prompt Learning} & &  & & & \\
    ~~DPL \cite{zhang2023domain} & 84.3 $\pm$ 0.4 & 97.3 $\pm$ 0.2 & 84.2 $\pm$ 0.2 & 52.6 $\pm$ 0.6 & 79.6 \\
    \rowcolor{mygray} ~~DAMP (Ours) & \textbf{84.5 $\pm$ 0.3} &  \textbf{97.4 $\pm$ 0.2} & \textbf{85.0 $\pm$ 0.4} & \textbf{53.7 $\pm$ 0.2} & \textbf{80.2} \\
    \bottomrule
    \end{tabular}}
    \label{tab:main_results_dg}
    \end{center}
    \vspace{-10pt}
    \end{table*}

We show that with only minor changes, our method can also be used for domain generalization tasks.

\textbf{Datasets.} We use four popular DG datasets in this experiment, namely, VLCS \cite{fang2013unbiased}, PACS \cite{li2017deeper}, Office-Home \cite{venkateswara2017deep} and TerraIncognita \cite{beery2018recognition}. {\bf VLCS} contains images from PASCAL VOC 2007 (V), LabelMe (L), Caltech (C) and SUN (S). There are 5 object categories shared by all domains: bird, car, chair, dog and person. {\bf PACS} collects totally 9,991 images from Photo (P), Art painting (A), Cartoon (C) and Sketch (S) with 7 common categories. {\bf Office-Home} is originally used in domain adaptation, which contains images from Art (Ar), Clipart (Cl), Product (Pr) and Real-World (Rw) across 65 categories. {\bf TerraIncognita} comprises a collection of wildlife photographs captured by cameras at various locations. We follow \cite{gulrajani2020search} to use 4 locations, i.e., \{L38, L43, L46, L100\}, for the DG task, which have totally 24,788 samples of 5 classes. 

\textbf{Experimental Setup.} For the DG task, we implement our method on DomainBed \footnote{https://github.com/facebookresearch/DomainBed}, a standard DG benchmark in the community. We strictly follow \cite{gulrajani2020search} to split each domain into 80\% training data and 20\% validation data, and use standard training-domain validation for model selection. The results are obtained by three trials with seed=\{1,2,3\}.

Different from domain adaptation, in DG we cannot access any target sample during training. Therefore, we remove $\mathcal{L}_{im}$ and target-related terms in $\mathcal{L}_{sup}$, $\mathcal{L}_{sc}$ and $\mathcal{L}_{idc}$ for optimization. In this scenario, our method aims to elicit domain-invariant visual embeddings from multiple source domains and instance-compatible text embeddings for classification. Due to the absence of the target domain, large-scale pre-trained knowledge becomes more important in DG. Therefore, CLIP-based methods will have significantly better results than traditional ones. For fair comparison, all compared baselines are built on CLIP, and we use the ViT-B/16 vision backbone for all datasets following \cite{zhang2023domain}. 

Specifically, there are three categories of baselines for comparison. The first category of methods fine-tune the image encoder of CLIP using common DG algorithms (e.g., like ERM, DANN) and freeze the text encoder for classification. The second category directly use the zero-shot ability of CLIP and prompt the text encoder with manually designed prompts (`a photo of [CLS]') for classification. The third category resorts to learnable prompts for adapt the pre-trained CLIP to specific domains. For our method, the hyperparameter configuration is consistent with the ones in UDA and MSDA.

\textbf{Experimental Results.} We report the mean accuracies as well as standard derivation on four datasets in Table \ref{tab:main_results_dg}. Our DAMP achieves the best average accuracy of 80.2\% across all datasets. Compared to fine-tuning-based methods like ERM, CORAL and DANN, DAMP brings significant improvements of 4.3\%, 3.3\% and 3.2\% respectively in average accuracy. We conjecture the reason is that large-scale pre-training is a very effective approach to bridge the domain gap. However, fine-tuning the image encoder is prone to destroy the pre-trained knowledge encoded in CLIP. This demonstrates the superiority of adapting pre-trained models via prompt learning over fine-tuning for domain generalization. On the other hand, DAMP also surpasses the vanilla zero-shot CLIP model by 6.5\%, showing the benefits of learning adaptive prompts compared to relying solely on pre-trained knowledge. Compared with DPL that only prompts the text encoder, our method prompts both vision and textual modalities for generalizing both visual images and textual semantics to unseen domains. Besides, two regularizations (i.e., $\mathcal{L}_{idc}$ and $\mathcal{L}_{sc}$) encourage the embeddings to be more domain-agnostic, thus outperforming DPL on all datasets. Even though the margins appear small, i.e., 0.6\% over DPL, it is well-recognized that further advancing the state-of-the-art on DG benchmarks is extremely challenging. Even slight gains of 1\% are considered significant and difficult in the DG community, which typically indicate non-trivial improvements in the robustness and generalization abilities of the model across diverse domains.

\section{Additional Analytical Experiments for UDA} 
\label{sec:appendix_analytical}
\subsection{Confusion Matrix Visualization} 

\begin{figure*}[t]
    \centering
     \includegraphics[width=\textwidth]{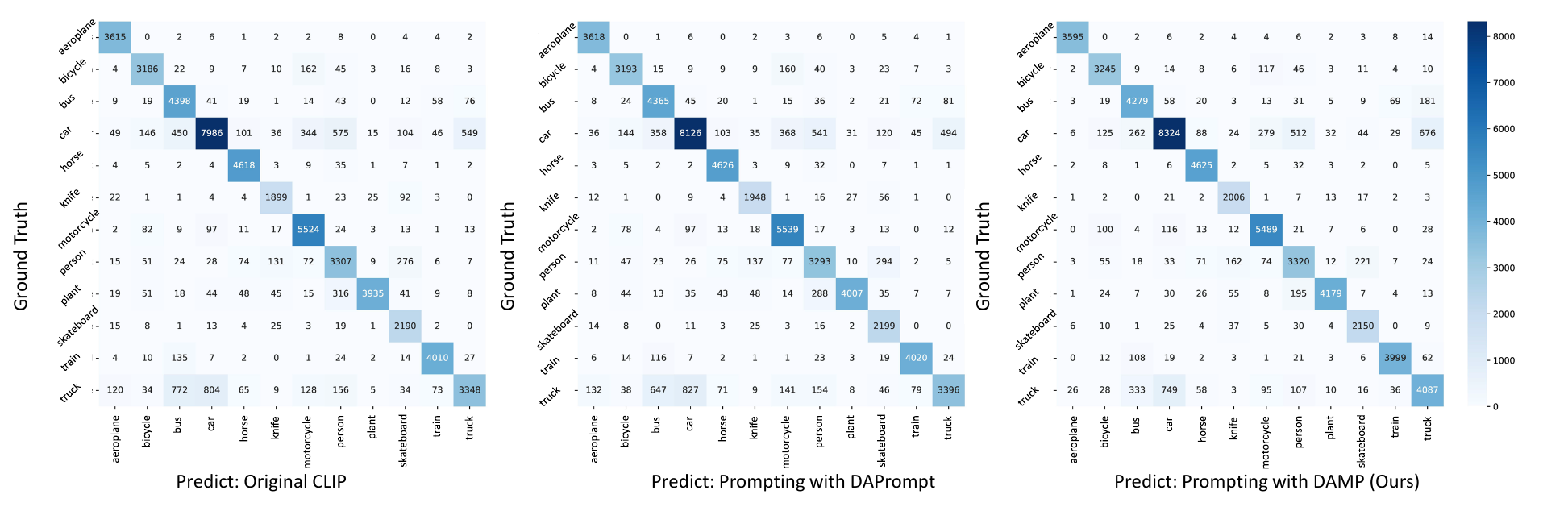}
     \vspace{-4pt}
     \caption{Visualization of confusion matrixes yield by different methods on VisDA-17 dataset with ViT-B/16 backbone.}
     \label{fig:confusion_matrix}
  \vspace{-14pt}
  \end{figure*}

To illustrate how our method benefits UDA, we visualize the confusion matrixes obtained by different methods in Fig. \ref{fig:confusion_matrix}. We can observe that directly using the zero-shot classification capability of CLIP can easily lead to confusion between categories. For instance, the model may easily predict "car" as "bus" or "truck," or predict "bicycle" as "motorcycle", because these categories are conceptually similar. In contrast, DAPrompt adjusts the textual semantics of categories specifically for each domain (dataset), making it clearer to distinguish the semantic differences between categories and to some extent alleviating the confusion problem. However, DAPrompt does not perform any adaptation in the visual modality, making it susceptible to domain shifts in the visual modality. Additionally, DAPrompt uses class-level semantic embeddings for classification, which does not take into account variations within categories. Imagine if the visual embedding of a truck in the multimodal space is closer to the semantic representation of "bus" than "truck." In this case, DAPrompt would have no way of classifying it as a truck. In contrast, our method allows the semantic embeddings to dynamically adjust their positions based on visual cues for each sample, providing a customized set of semantic embeddings for classification, and therefore performs better in disambiguation compared to CLIP and DAPrompt.

\subsection{Effectiveness of the Pseudo-label Strategy}

\begin{figure*}[t] 
    \centering 
    \subfloat[Ar$\rightarrow$Cl]{
    \includegraphics[width=0.23\linewidth] {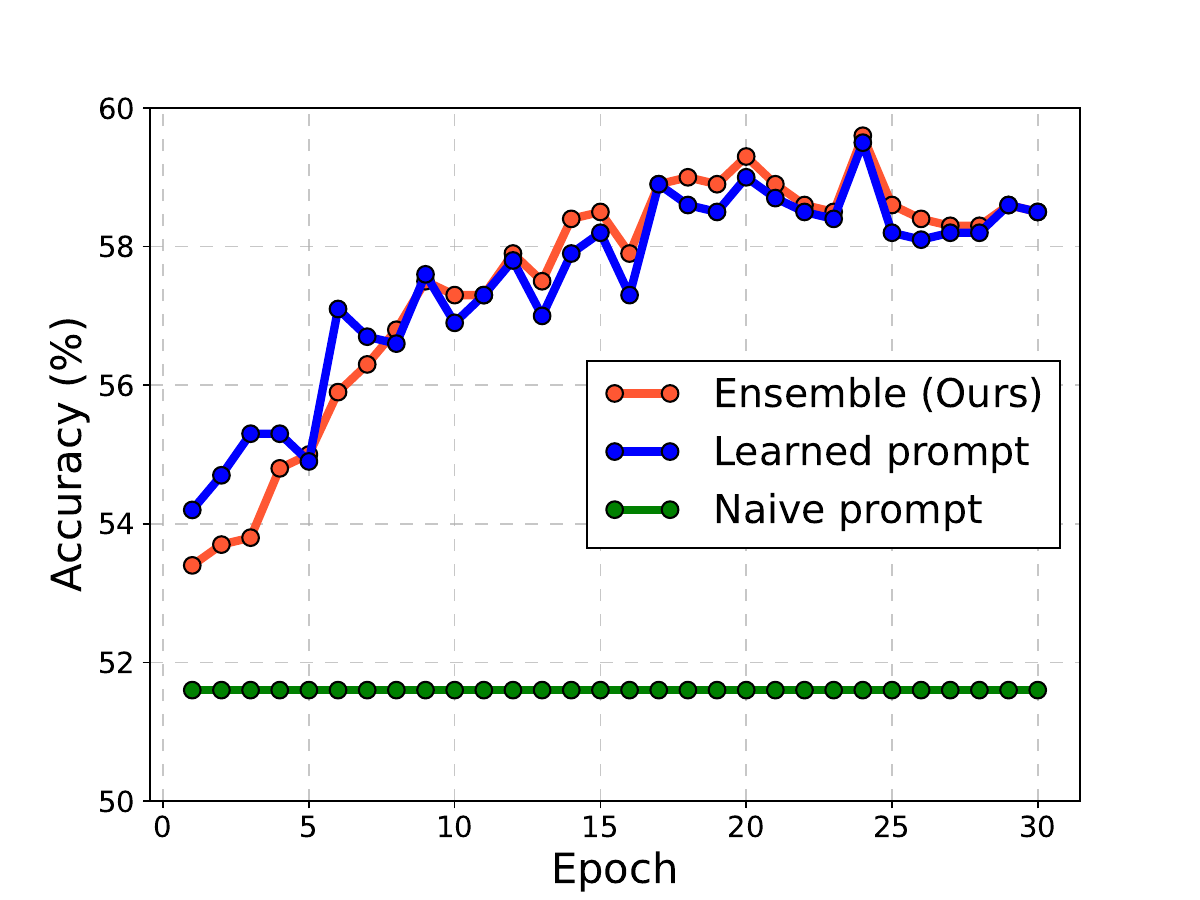} 
    }
    \subfloat[Ar$\rightarrow$Pr]{
    \includegraphics[width=0.23\linewidth] {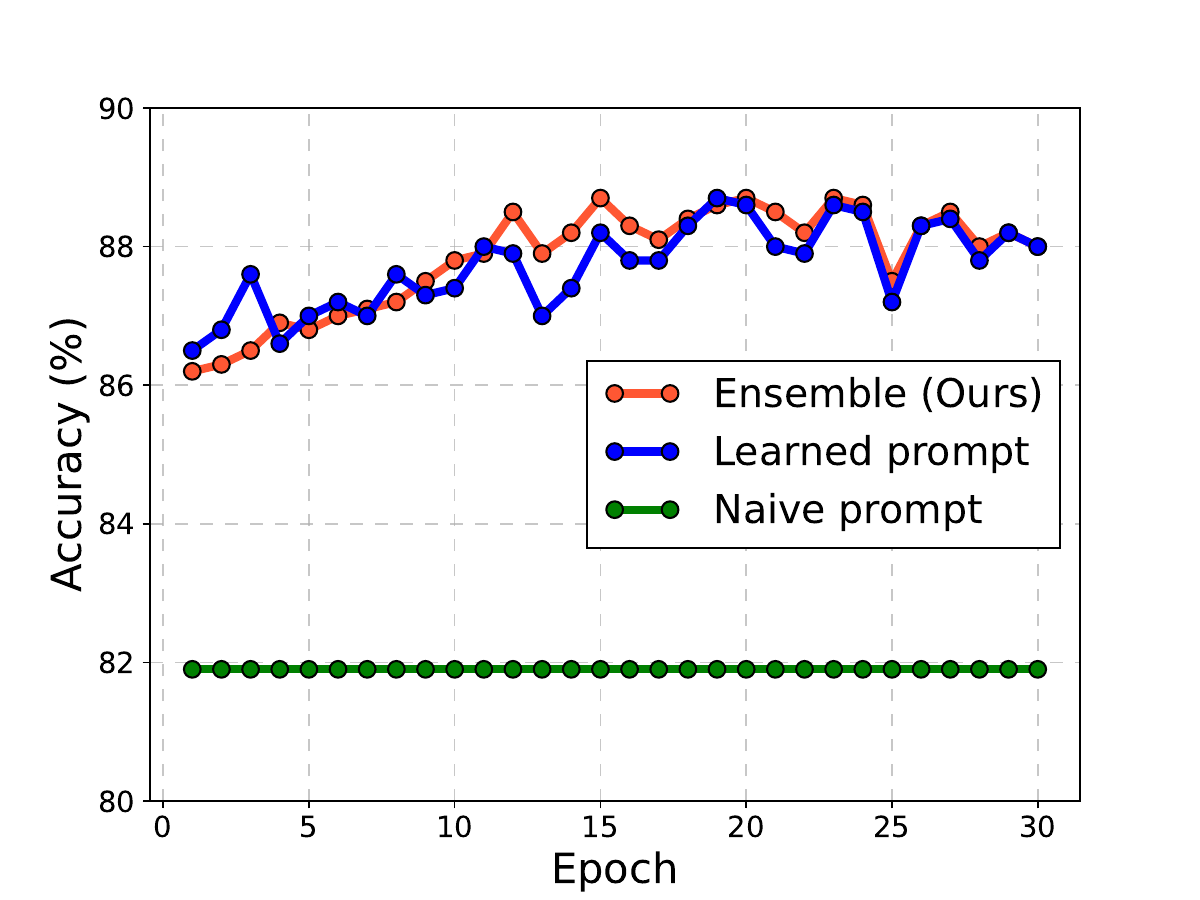}
    }
    \subfloat[Pr$\rightarrow$Ar]{
    \includegraphics[width=0.23\linewidth] {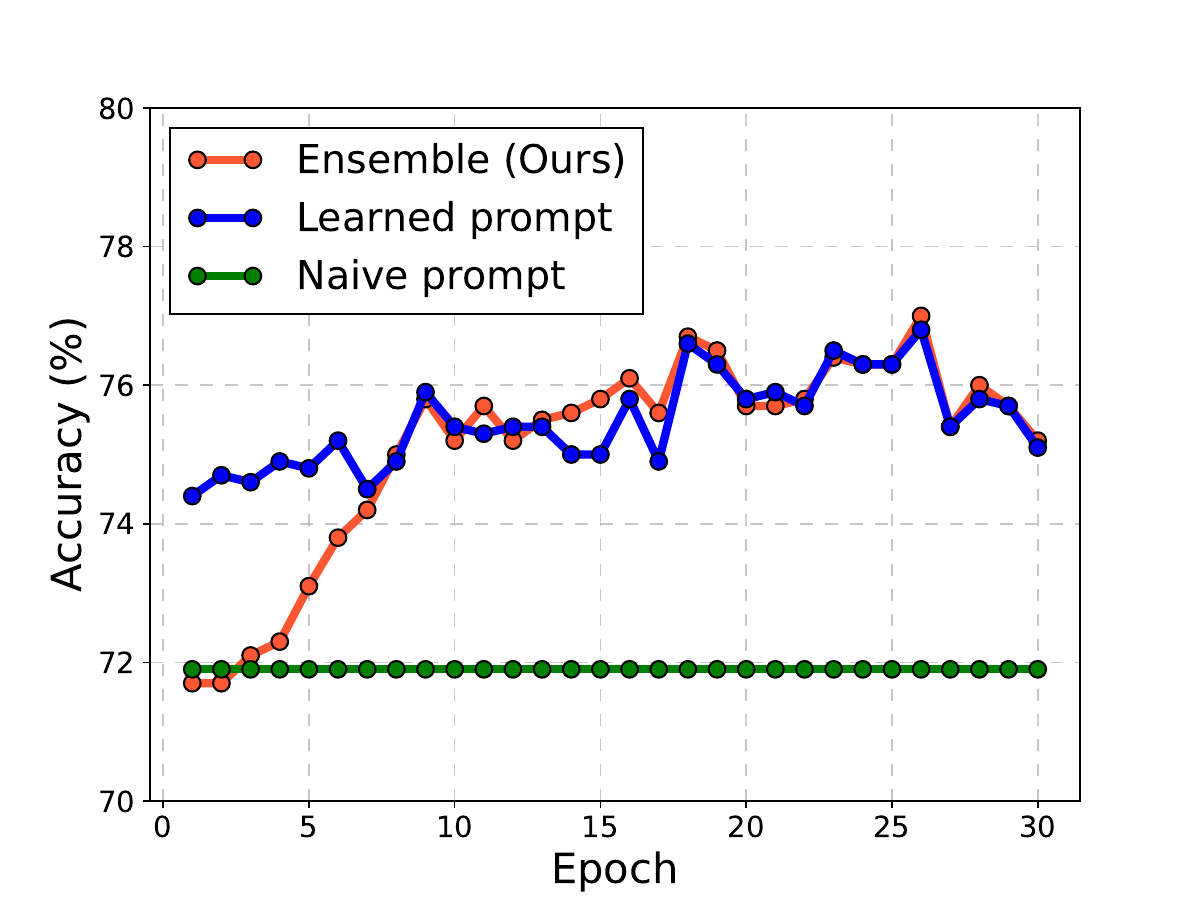}
    }
    \subfloat[Pr$\rightarrow$Cl]{
    \includegraphics[width=0.23\linewidth] {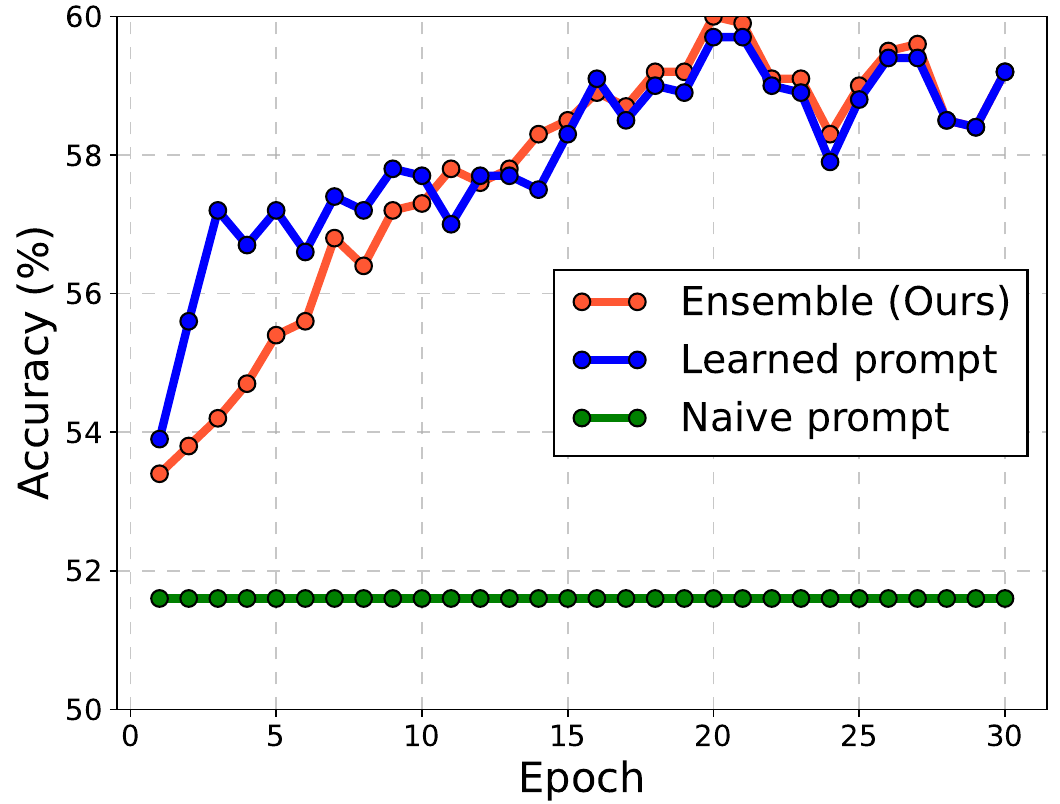}
    }
    \caption{Pseudo-label accuracies of different strategies. We choose four tasks on Office-Home as examples (ResNet-50).} 
    \label{fig:pseudo_label_acc} 
    \vspace{-10pt} 
  \end{figure*}

  \begin{table*}[t]
    \centering
    \caption{Classification accuracies of our method with (w/) and without (w/o) Parameter-Sharing (PS) strategy in the mutual prompting module. The results are obtained on {\bf Office-Home} dataset.}
    \vspace{-5pt}
    \resizebox{\linewidth}{!}{
    \begin{tabular}{lccccccccccccccc}
    \toprule
       {\textbf{Method}} & $f_v$ & {Ar$\rightarrow$Cl} & {Ar$\rightarrow$Pr} & {Ar$\rightarrow$Rw} & {Cl$\rightarrow$Ar} & {Cl$\rightarrow$Pr} & {Cl$\rightarrow$Rw} & {Pr$\rightarrow$Ar} & {Pr$\rightarrow$Cl} & {Pr$\rightarrow$Rw} & {Rw$\rightarrow$Ar} & {Rw$\rightarrow$Cl} & {Rw$\rightarrow$Pr} & {\textbf{Avg.}} \\ 
        \midrule
        DAMP w/o PS  & \multirowcell{2}{ResNet-50} & 59.5	&88.6&	86.4&	76.5&	89.0&	86.8&	76.6&	59.8&	86.9&	77.1&	60.4&	89.5&	78.1    \\      

        DAMP w/ PS & & 59.7 & 88.5 & 86.8 & 76.6 & 88.9 & 87.0 & 76.3 & 59.6 & 87.1 & 77.0 & 61.0 & 89.9 & 78.2 \\
\midrule
\midrule    

        DAMP w/o PS & \multirowcell{2}{ViT-B/16} & 75.9 &	93.7&	92.2&	86.1&	94.3&	91.7&	85.7&	76.1&	92.0&	85.5&	76.1&	93.6&	86.9    \\   

        DAMP w/ PS & & 75.7 & 94.2 & 92.0 & 86.3 & 94.2 & 91.9 & 86.2 & 76.3 & 92.4 & 86.1 & 75.6 & 94.0 & 87.1 \\ 
    \bottomrule
    \end{tabular}}
    \label{tab:parameter_sharing}
\end{table*}

To evaluate the effectiveness of our ensemble-based pseudo-label strategy described in Appendix. \ref{sec:appendix_Implementation}, we conduct experiments to compare with other two pseudo-label strategies, i.e., using the naive prompts (`a photo of a [CLS]') for zero-shot prediction and using the learned prompts $\bm{p}_{1:N}$ with {\it post-model} multual prompting. The former strategy is used in DAPrompt \cite{ge2022domain} and the latter is the outputs of our model. As shown in Fig. \ref{fig:pseudo_label_acc}, we can observe a common phenomenon across all tasks. Initially, due to the zero-shot capability of CLIP, it can provide high-quality pseudo-labels for the target domain samples, resulting in high accuracy achieved in the first epoch for learned prompts. However, as the proportion of zero-shot pseudo-labels is still high at this point, the accuracy of the ensemble pseudo-labels remains lower than that of the learned prompt-based pseudo-labels. As training progresses (after 10 epochs), the ensemble pseudo-labels outperforms the other two strategies. We conjecture the reason is that the incorporation of certain prior knowledge (naive prompts) helps alleviate the risk of overfitting the learned prompts to the source domain. This enables the model to achieve accuracy that cannot be attained by relying solely on learned prompts for pseudo-labeling. Finally, the accuracy of the ensemble gradually converges towards the accuracy of the learned prompts.

\subsection{Effectiveness of Parameter-Sharing Strategy} 
Table \ref{tab:parameter_sharing} shows the impact of parameter-sharing $G$ on the performance of our method. It turns out that the parameter-sharing strategy (w/ PS) slightly outperforms the version without parameter-sharing (w/o PS) on both vision backbones. Parameter-sharing enables a single prompting module $G$ to transform both visual and textual embeddings bidirectionally. This allows richer cross-modal interactions and fusion between the modalities, eliciting better domain- and modality-shared representations. In contrast, without parameter-sharing, the promptings of vision of vision and text modalities will be more independent, and more tunable parameters make it difficult to train and less effective. The consistently positive gains across various backbones indicate that parameter-sharing is an effective and generalizable design choice for mutual prompting in DAMP. It enables a single compact module to prompt both modalities flexibly for domain adaptation. 

\end{document}